\documentclass[10pt,twocolumn,letterpaper]{article}

\usepackage[T1]{fontenc}
\usepackage[pagenumbers]{cvpr} 

\usepackage{multirow}
\usepackage{booktabs}
\usepackage{multirow}
\usepackage{xcolor}
\usepackage{makecell}

\usepackage{graphicx}
\usepackage{caption}
\usepackage{subcaption}

\definecolor{cvprblue}{rgb}{0.21,0.49,0.74}
\usepackage[pagebackref,breaklinks,colorlinks,allcolors=cvprblue]{hyperref}

\newcommand{\benchname}{TechImage-Bench\xspace}

\def\paperID{7329} 
\def\confName{CVPR}
\def\confYear{2026}

\title{\benchname: Rubric‑Based Evaluation for Technical Image Generation}

\author{Minheng Ni$^{1,2~\dagger}$, Zhengyuan Yang$^{3~\dagger}$, Yaowen Zhang$^{2~\dagger}$, Linjie Li$^3$, Chung-Ching Lin$^3$, Kevin Lin$^3$, \\
Zhendong Wang$^3$, Xiaofei Wang$^3$, Shujie Liu$^3$, Lei Zhang$^1$, Wangmeng Zuo$^{2}$, Lijuan Wang$^3$\\
{\small $^1$Hong Kong Polytechnic University \quad $^2$Harbin Institute of Technology \quad $^3$Microsoft}\\
{\href{https://kodenii.github.io/ProImage-Bench-web}{Project Website}}
}

\begin{document}
\twocolumn
[{%
\renewcommand\twocolumn[1][]{#1}%
\maketitle
\begin{center}
    \centering
    \vspace{-6mm}
    \captionsetup{type=figure}
\includegraphics[width=\linewidth,height=8cm]{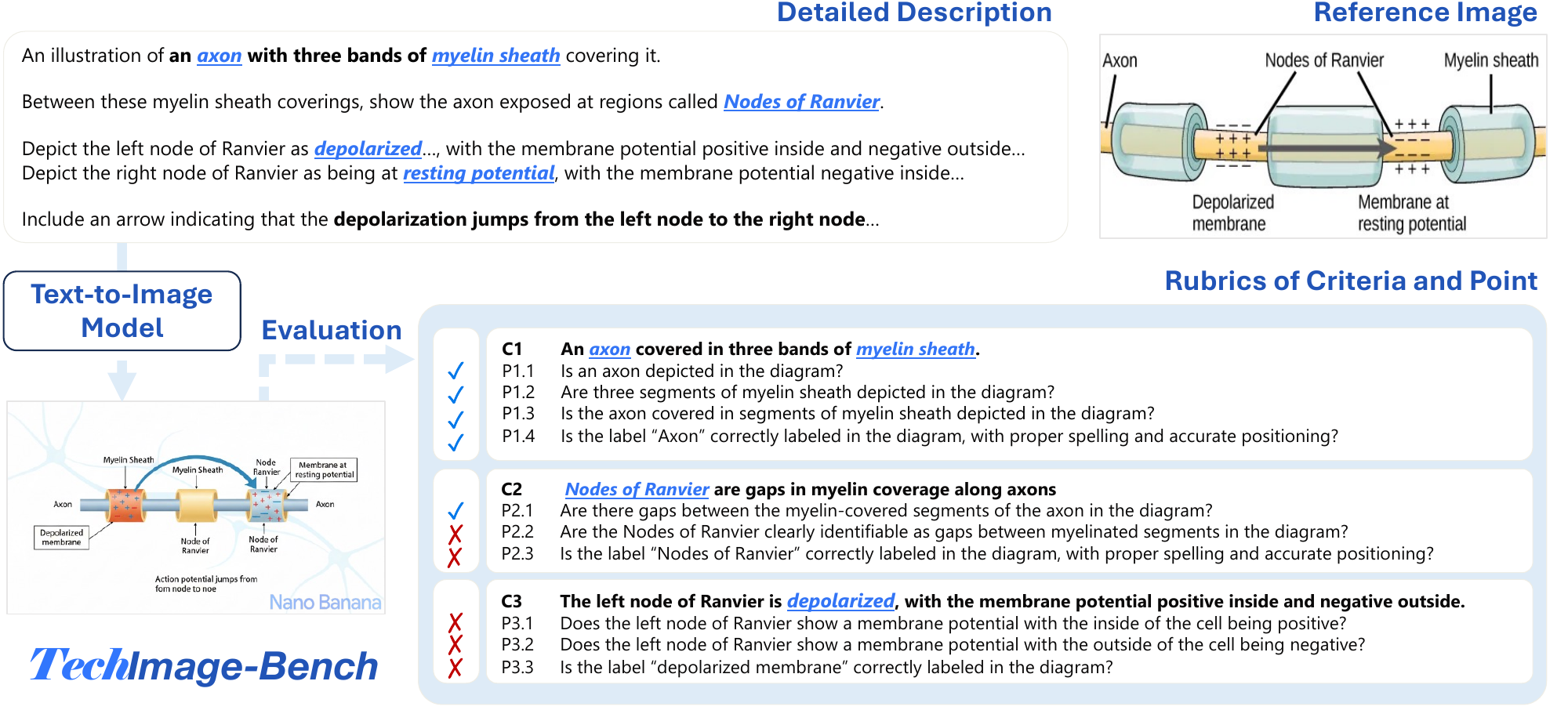}
    \captionof{figure}{\textbf{We propose \benchname, a rubric‑based evaluation benchmark for technical image generation.} The benchmark will be publicly available.} 
    \label{fig:teaser}
\end{center}%
}]

\begin{abstract}

We study technical image generation, where a model must synthesize information‑dense, scientifically precise illustrations from detailed descriptions rather than merely produce visually plausible pictures. To quantify the progress, we introduce \benchname, a rubric-based benchmark that targets biology schematics, engineering/patent drawings, and general technical illustrations. For 654 figures collected from real textbooks and technical reports, we construct detailed image instructions and a hierarchy of rubrics that decompose correctness into 6,076 criteria and 44,131 binary checks. Rubrics are derived from surrounding text and reference figures using large multimodal models, and are evaluated by an automated LMM-based judge with a principled penalty scheme that aggregates sub-question outcomes into interpretable criterion scores. We benchmark several representative text-to-image models on \benchname and find that, despite strong open-domain performance, the best base model reaches only 0.801 rubric accuracy and 0.576 criterion score overall, revealing substantial gaps in fine-grained scientific fidelity. Finally, we show that the same rubrics provide actionable supervision: feeding failed checks back into an editing model for iterative refinement boosts a strong generator from 0.660 to 0.865 in rubric accuracy and from 0.382 to 0.697 in criterion score. \benchname thus offers both a rigorous diagnostic for technical image generation and a scalable signal for improving specification-faithful scientific illustrations.

\end{abstract}

\section{Introduction}
\label{sec:intro}

Recent text‑to‑image (T2I) models produce photo‑realistic and semantically aligned images for open‑domain prompts ~\citep{wan2025wan,zheng2024cogview3,seedream2025seedream}. The strong performance motivates real utility for technical image generation, \eg, scientific illustrations, engineering drawings, and patent‑style diagrams~\citep{kolpakov2006biouml,gowaikar2024agentic,tang2024exploring,fang2017creatism,elasri2022image}. In this study, we aim to define and benchmark the technical image generation problem, where the model takes technical documentation, and produces an image that is not only visually plausible but also {information‑dense} and {factually precise}. Representative scenarios include biology schematics, engineering or patent embodiments, and scientific diagrams. However, this task requires evaluation that goes beyond aesthetics to verifiable correctness.

Different from daily images, fine‑grained mistakes common in technical figures, such as misplaced organelles, omitted mechanical components, or incorrect data relations, can invalidate the entire illustration while remaining subtle to non‑experts and invisible to generic image metrics\citep{li2024aigiqa,min2025exploring,agnolucci2024arniqa} or "few‑second glance" human judgments. As shown in Figure~\ref{fig:teaser}, the state-of-the-art proprietary models generate seemingly correct but far from usable technical images. We argue that the existing holistic scoring is too coarse to capture these failure modes reliably. The key to making consistent improvement for the technical image generation task is by creating a consistent and comprehensive grading signal across different domains, tasks, and samples.

Following insights from rubric‑driven large language model (LLM) evaluations, \eg, PaperBench~\citep{starace2025paperbench}, HealthBench~\citep{arora2025healthbench}, we argue that assessment should be decomposed into explicit rubrics instead of single holistic scores such as CLIP score~\citep{hessel2021clipscore}, aesthetic~\citep{murray2012ava}, human preference score~\citep{wu2023human}, or large multimodal model (LMM) grading in this special problem. Therefore, we propose to separate task checks from domain ones. Specifically, this corresponds to task-specific rubrics grounded in each sample's given input documentation, and domain-specific rubrics shared among samples within each task, capturing each domain and task's unique requirements. Converting each abstract criterion into binary, unit‑test‑like questions yields scores that are more reliable, interpretable, and actionable than holistic ratings, while remaining scalable with LMM‑based judging. Our analyses also validate that specifying the aspect to judge based on the rubrics significantly simplifies the task for the LMM grader and yields better alignment with human experts, even with weaker LMMs.

Building on this view, we introduce TechImage‑Bench, a breakdown benchmark for technical image generation. The core task of this benchmark is to generate precise technical illustrations based on detailed textual descriptions, \eg, captions, figure titles, or research report excerpts. We extract text–image pairs from authentic professional textbooks and reports, including captions, titles, and in-text figure references, and use an LMM to extract implicit constraints from these sources. These constraints are then transformed into a series of objective, binary (yes/no) conditions, forming our dataset's detailed rubrics, which decompose the evaluation into independent sub-criteria across two dimensions: task and domain.
An automated LMM Evaluator answers each prompt independently and aggregates at the criterion level using a principled penalty scheme. The current benchmark covers Biology, Engineering, and General  illustrations, totaling 654 tasks, 6,076 criteria, and 44,131 binary checks, and shows the stability of the evaluation.

We evaluate several state-of-the-art T2I models on \benchname, including GPT-Image~\citep{hurst2024gpt} and Nano Banana 2~\citep{Nano_Banana_2}. Experimental results clearly show that although these models excel in general tasks, their average reproduction accuracy on our benchmark remains lower than 80\%, with only Nano Banana 2 barely exceeding this threshold. This finding quantitatively reveals severe deficiencies in current SOTA models' fine-grained accuracy. Importantly, rankings do not perfectly align with perceived general‑domain quality, underscoring that progress on professional and high‑valued images requires methods that explicitly reason about and satisfy specifications. Furthermore, to validate the effectiveness of our rubric-based evaluation, we conduct a rule-based editing experiment: low-scoring rubrics are fed back into an editing model, which iteratively refines the generated image. The revised images achieve significantly higher scores on \benchname, confirming the practical value of our rubric decomposition in guiding the improvement in technical image generation, either as offline grading scores or online reward values. 

Our contributions are threefold:
\begin{itemize}
\setlength{\emergencystretch}{2em}
    \item We formalize the technical image generation problem and propose a rubric‑based evaluation that decomposes verification into task‑specific and domain‑specific checks.
    \item We collect and release \benchname, spanning 654 tasks / 6,076 criteria / 44,131 binary checks across biology, engineering, and general domains, with an automated LMM Evaluator and principled scoring. 
    \item We provide comprehensive analyses showing (i) misalignment between open‑domain prowess and professional accuracy, and (ii) significant gains from rubric‑guided refinement, charting a path that produces specification‑faithful and professional‑grade images.
\end{itemize}

We released the \benchname code and dataset to encourage future research on improving the model's capability for technical image generation.
\section{Related Work}
\paragraph{Text-to-image Generation}
Recent advances in diffusion and transformer-based models have greatly improved open-domain text-to-image synthesis~\citep{zhang2023adding,ruiz2023dreambooth,saharia2022photorealistic,Gemini2_5_Flash}. Systems like DALL-E~\citep{ramesh2021zero} and Stable Diffusion~\citep{rombach2022high} can produce high-quality images from natural language prompts, demonstrating unprecedented diversity and fidelity. Google's Imagen model~\citep{saharia2022photorealistic} further showed photorealistic generation capabilities, underscoring the rapid progress in this field. These powerful generative models have, in turn, motivated attempts to apply them in high-value specialized domains such as scientific and technical illustration~\citep{rodriguez2023ocr,gillani2025textpixs,mondal2024scidoc2diagrammer,wang2025magicgeo,hnatkowska2021activity}. Their strong general capabilities suggest potential for automating tasks like diagram creation or academic figure drawing, which are traditionally labor-intensive~\citep{wei2025words,zhong2024ai,bamouh2025towards,dias2023automated,malik2025mathematikz}. However, most such uses to date have been ad hoc explorations by individual users. The community has lacked rigorous benchmarks to track progress on these applications, given the challenges of the required detailed perception and domain knowledge.

\paragraph{Technical Image Generation}
A growing line of work targets the generation of domain-specific diagrams and scientific images from text descriptions~\citep{lee2025text,xing2025empowering,belouadi2023automatikz,belouadi2024detikzify,xiao2025uml}. Early efforts include code-based approaches~\citep{rodriguez2023starvector,carlier2020deepsvg,xing2024svgdreamer,jain2023vectorfusion}, where LLMs can be prompted to emit diagram specification code, \eg, TikZ or SVG scripts, that a renderer converts into an image. For example, AutomaTikZ~\citep{belouadi2023automatikz} and DeTikZify~\citep{belouadi2024detikzify} demonstrated text-guided generation of scientific graphics by producing TikZ code. These methods achieve element-level precision but involve a cumbersome write–compile–review loop, limiting their practicality for complex figures. Others have explored direct image-based generation. FigGen~\citep{rodriguez2023figgen} introduced the text-to-scientific figure task using diffusion models, highlighting challenges in going beyond natural images. More recently, dedicated benchmarks and datasets have begun to emerge. SridBench~\citep{chang2025sridbench} is the first benchmark focused on scientific research figure synthesis, compiling 1,120 illustrations from papers across 13 disciplines. Two very recent efforts further advance this area. VisBench~\citep{heiland2001visbench} presents a seven-criterion evaluation suite assessing content, layout, visual clarity, and user interactivity of generated figures. The InteractScience benchmark~\citep{chen2025interactscience} focuses on interactive scientific content generation, which evaluates LLMs that produce interactive scientific demonstrations by combining rigorous functional tests with visually-grounded checks against reference outputs. Compared to these works, \benchname goes further in both scope and rigor: it spans multiple technical domains and employs a rubric-based, hierarchical evaluation that decomposes each figure's requirements into fine-grained criteria and binary verification checks.

\paragraph{Text-to-Image Evaluation}
Text-to-Image (T2I) generation is typically evaluated using holistic metrics, such as Fréchet Inception Distance (FID)\citep{heusel2017gans}, Inception Score (IS)\citep{salimans2016improved}, or CLIP Score~\citep{hessel2021clipscore}. In recent years, evaluation paradigms have gradually shifted toward prompt-guided visual question answering methods, such as VQA Score~\citep{hu2023tifa}. However, scientific and technical illustrations often require strict structural correctness, domain-specific visual logic, and precise data representation, which general-purpose methods struggle to reliably assess. To address this gap, our work shifts T2I evaluation toward expert-reviewed, sample-specific verification, ensuring that the generated content meets the stringent requirements of professional applications.

\paragraph{Evaluation of Hard-to-Verify Tasks}
The NLP community has developed methods to evaluate complex, open-ended outputs where simple ground-truth matching is inadequate. A prominent example is PaperBench~\citep{starace2025paperbench}, which assesses an AI agent's ability to replicate research papers. PaperBench uses a manually crafted hierarchical rubric to break down each replication into sub-tasks with clear grading criteria, yielding over 8,000 individually gradable items. Crucially, it employs an automated judge, \ie, an LLM, to grade each sub-task against the rubric, achieving scalable evaluation aligned with expert judgment. In the medical domain, HealthBench~\citep{arora2025healthbench} introduced a rubric-driven evaluation of LLM responses to clinical scenarios. These benchmarks underscore the importance of multi-dimensional, rubric-based evaluation for tasks where quality is complex and a single ground-truth answer is absent. Inspired by these approaches, \benchname adapts the rubric-driven methodology to visual content generation. We design hierarchical rubrics that capture the essential qualities of technical images, \eg, content accuracy, visual clarity, adherence to style conventions, and employ automated scoring procedures to ensure scalable yet principled evaluation.
\section{\benchname}

\begin{figure*}[h!]
    \centering
    \includegraphics[width=1.0\textwidth]{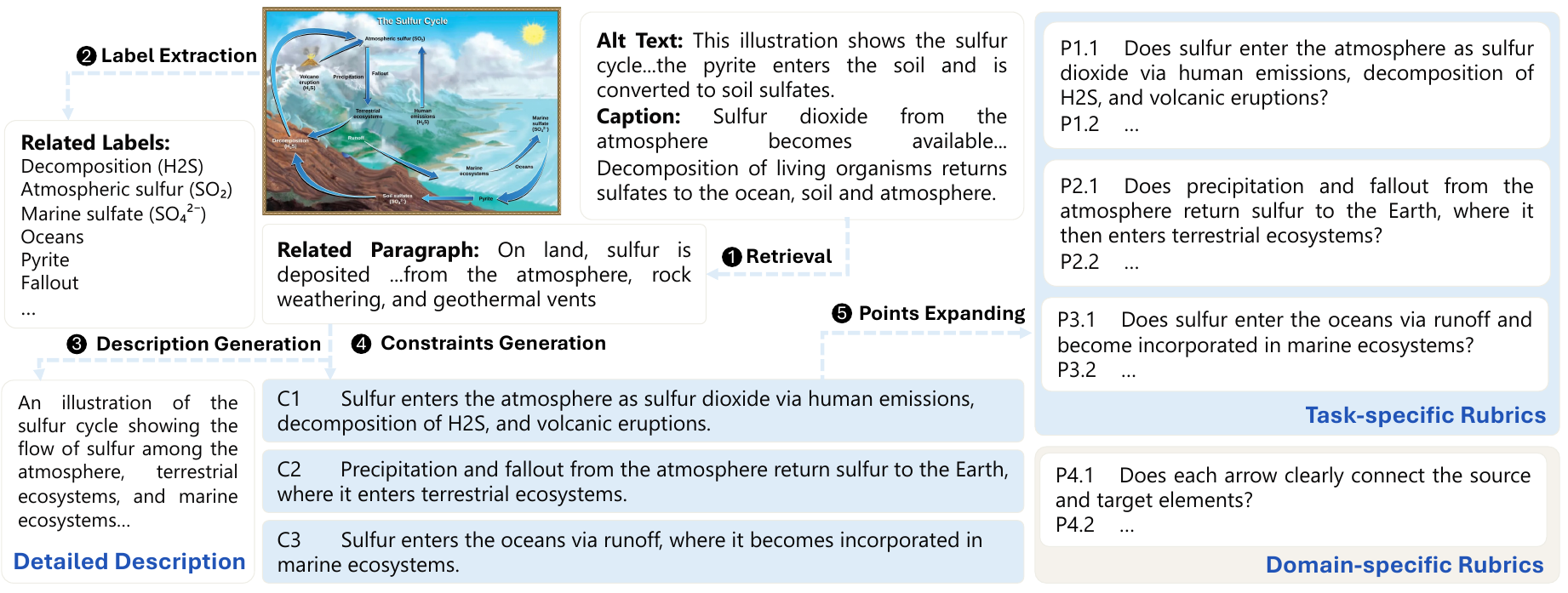}
    \caption{\textbf{The process of constructing detailed descriptions and rubrics.} 
(1) Retrieval: The alt text, the caption, and the related paragraph associated with the image are retrieved from the textbook.
(2) Label Extraction: Related labels are extracted from the image, \eg, decomposition, atmospheric sulfur, and marine sulfate. 
(3) Description Generation: A coherent and detailed image description is generated based on the retrieved text and related labels. 
(4) Constraint Generation: LMM identifies and extracts the implicit constraints encoded in the textual description. 
(5) Points Expanding: Each constraint is expanded into fine-grained binary scoring points, covering both inclusion and exclusion criteria.}
    \label{fig:method}
\end{figure*}

Although current generative models are remarkably powerful, they still tend to produce fine-grained errors when applied in professional scientific scenarios. These errors are often difficult for non-experts to discern, and existing evaluation methods struggle to capture the subtle details that are critical to scientific rigor. Just as we require unit tests to validate code, there is an urgent need for a reliable method to assess such hard-to-verify visual generation content. This task demands that models not only ensure semantic and visual alignment with the input description, but more crucially, guarantee fine-grained scientific accuracy.

\begin{figure*}[hpt!]
    \centering
    \includegraphics[width=\textwidth]{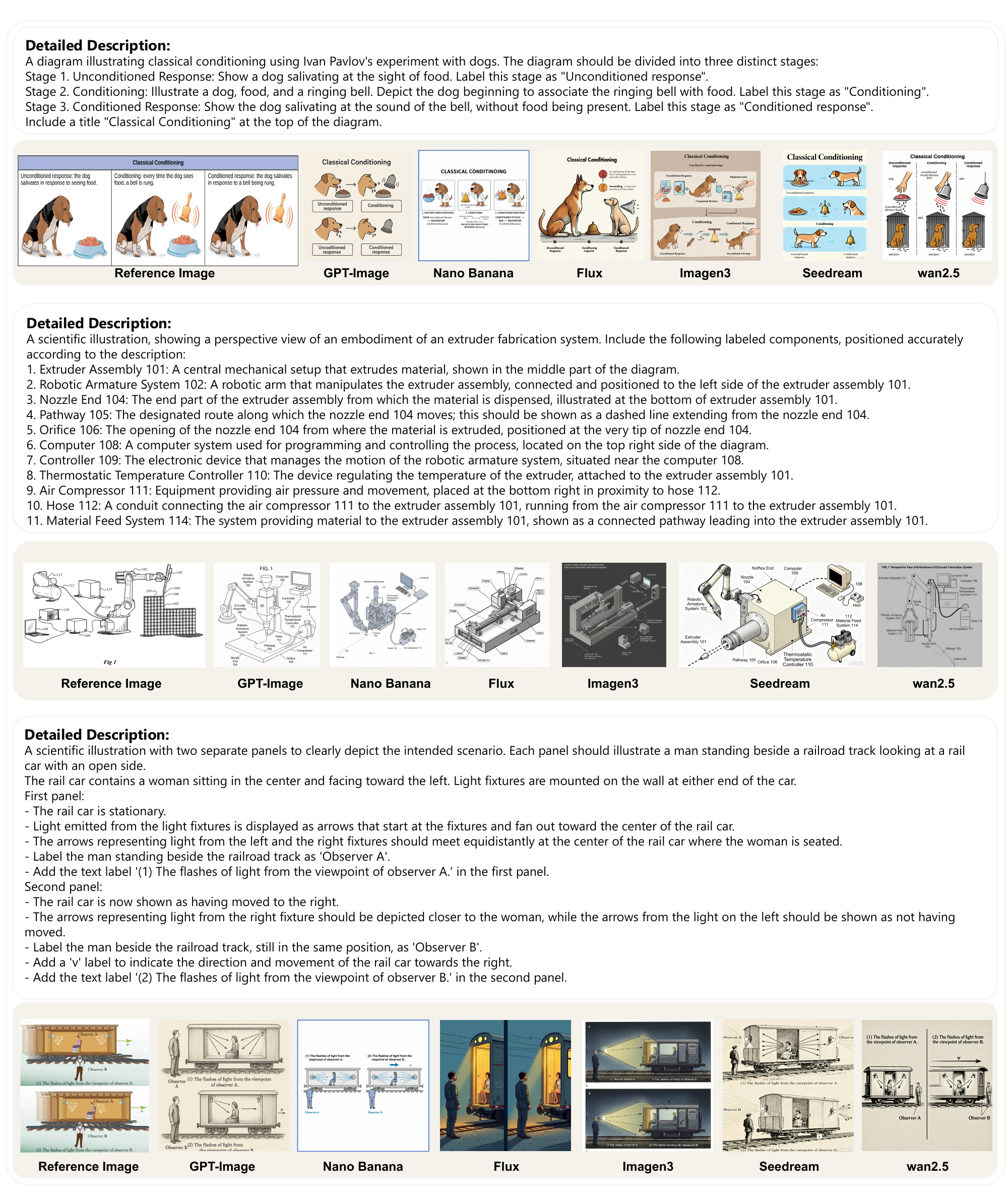}
    \caption{\textbf{Samples of dataset and generated images.}
(1) In the case of the biology domain, GPT-Image, Flux, and Imagen3 failed to depict the three stages as required, while Nano Banana, Seedream, and Wan2.5 included all three stages, but the logical progression of each stage was not clearly conveyed. 
(2) In the case of the engineering domain, GPT-Image, Nano Banana, Seedream, and Wan2.5 exhibit issues with missing labels and incomplete depiction of components as described in the detailed description, while Flux and Imagen3 produce images that do not align with the detailed description.
(3) In the case of the general domain, Flux did not follow the detailed description in its depiction, while the results from the other models contained issues with incorrect positioning of the light fixtures and improper direction of the light rays.
}
\label{fig:dataset}
\end{figure*}

\subsection{Problem Definition}

Consider a technical image instruction $t$ and a set of evaluation criteria $\mathcal{C} = \{c_1, c_2, \ldots\}$, where each $c_i$ represents an abstract evaluation item describing an independent and complete evaluation rule, \eg, ``the phospholipid bilayer is correctly depicted.'' We observe that in technical image generation, any evaluation rule is rarely entirely violated or fully satisfied, as shown in Figure \ref{fig:teaser}. It is difficult to judge each rule using a simple binary decision (yes/no), which brings significant challenges for objective evaluation.

Therefore, each criterion is further decomposed into a series of scoring points, \ie, $c_i = \{p_{i,1}, p_{i,2}, \ldots\}$, where each scoring point corresponds to an objective binary question, such as ``Does the cell membrane exhibit a bilayer structure?'', ``Is the bilayer composed of hydrophobic and hydrophilic ends?'', ``Is the bilayer symmetric?'' and so on. In this way, we are able to perform a fine-grained evaluation of the generated image.

We define the core problem as generating a technical image $x$ from a detailed textual instruction $t$, and evaluating the image through this set of scoring criteria $\mathcal{C}$ to accurately capture the subtle differences that are essential for the rigor of technical images.

\subsection{Dataset Construction}

As shown in Figure \ref{fig:method}, we constructed our dataset through a series of automated steps combined with meticulous manual filtering and refinement. As listed in Table \ref{tab:dataset}, the dataset is divided into three domains, biology, engineering, and general, based on the scenario. In total, our dataset contains 654 unique generation tasks, along with 6,076 finely annotated criteria and 44,131 scoring points.

\paragraph{Raw Image-text Corpus Collection}

We collected more than 10,000 examples of paragraphs containing images and detailed descriptions from publicly available websites, including textbooks, professional books, web pages, and patent documents.
We extracted the corresponding captions, alt text, and contextual references from the paragraphs surrounding each image. After manual screening and filtering, we selected the images that met our criteria to serve as the reference images in the dataset, along with their associated context.

\paragraph{Detailed Description Generation}

The constraints of a technical image are not solely contained in the caption or the alt text, which primarily serve accessibility purposes; rather, many requirements are embedded within the paragraphs that reference the image. Therefore, we (1) retrieve related paragraphs based on the caption and alt text, and then (2) extract the related labels in the reference image. By integrating this multimodal information, we (3) employ a large multimodal model (LMM) to generate a coherent and semantically detailed description, which provides a reliable foundation for image generation.

\paragraph{Rubrics Generation}

Another major challenge is how to construct evaluation rubrics. To address this, we divide the rubrics into two categories: task-specific and domain-specific. Domain-specific rubrics mainly consist of manually designed general scoring prompts that assess the overall quality of images, such as arrows, labels, etc. Task-specific rubrics focus on evaluating the consistency between the image content and the textual context. To achieve this, (4) we utilize the multimodal context collected in the previous stages to generate explicit or implicit constraints, \ie, criteria, on the image, and then (5) refine these extracted constraints into a set of binary scoring questions that can be clearly answered with yes or no. Through this series of steps, we can construct a set of concrete evaluation criteria for each technical image. However, the initially generated criteria may still vary in quality and contain redundant prompts. Therefore, these preliminary criteria are cross-validated with the detailed description to ensure consistency, and then deduplicated, filtered, and text-refined by an LMM and human experts to produce the final dataset. 

\paragraph{Human Refinement}

To ensure the rigor and reliability of the dataset, we recruited domain experts to review and revise the automatically generated detailed descriptions and rubrics. This process guaranteed the validity of all generated content, allowing for cross-validation by independent experts. For additional details of implementation, refer to \textbf{Supplementary Materials}. 

\subsection{LMM Evaluator}

To enable large-scale automated evaluation, we develop an LMM‑based automatic evaluator. For each image, the evaluator sequentially responds to every scoring prompt. Each scoring prompt is submitted to the model individually to ensure that every question is answered accurately. The alt text and title of the image are also provided during evaluation to help the model better understand the intended meaning of the image and make precise judgments. For image evaluation, we design two metrics. Rubric accuracy measures the success rate of scoring points, and is defined as:
\begin{equation}
\textrm{Accuracy} = 1 - \frac{\sum e_i}{\sum |c_i|},
\end{equation}
and criterion score measures the degree of overall compliance with the constraints:
\begin{equation}
\textrm{Score} = \frac{1}{|\mathcal{C}|} \sum 0.5^{e_i},
\end{equation}
where $e_i$ denotes the number of failed points under the $i$-th criterion. If two errors fall in the same criterion versus different criteria, the rubric's accuracy, focusing on low-level, is the same in both cases, but the former will have a higher criterion score, focusing on high-level. For additional details of dataset, refer to \textbf{Supplementary Materials}. 

\section{Experiments}

\subsection{Experimental Setup}

We evaluated a series of state-of-the-art models, GPT-Image~\cite{GPT-4o}, GPT-Image-1.5~\cite{GPT-Image-1.5}, Nano Banana~\cite{Gemini2_5_Flash}, Nano Banana Pro~\cite{Nano_Banana_Pro}, Nano Banana 2~\cite{Nano_Banana_2}, Flux~\cite{labs2025flux}, Imagen-3~\cite{baldridge2024imagen}, Seedream~\cite{seedream2025seedream}, and Wan2.5~\cite{WAN2_5_T2I_PREVIEW}, on our TechImage-Bench. Specifically, following our detailed descriptions, we asked each model to generate images in professional styles, and then used OpenAI o4-mini as the LMM evaluator to assess the results. All models were given the same prompts and followed the same settings. For additional details of experimental setup, refer to \textbf{Supplementary Materials}. 

\subsection{Quantitative Results}

As shown in Table \ref{tab:overall}, from the overall results, although current models demonstrate strong generative capabilities on open-domain image tasks, there remains a significant gap in fine-grained scientific consistency. The best-performing model is Nano Banana 2, while GPT-Image and Nano Banana achieve overall accuracies of less than 70\%. This indicates that even state-of-the-art T2I systems still face substantial challenges when generating complex, structurally precise, and highly specialized technical images. More importantly, the overall score, which focuses on overall capability, remains below 0.6. This not only reflects the stringent requirements and fine-grained difficulty of TechImage-Bench in scientific rigor, but also highlights its strong discriminative power in differentiating model capabilities.

In addition, the difficulty levels across the three sub-tasks differ significantly, with engineering being the most challenging for all models. Even the best-performing model shows clearly lower results compared with the other two categories. This finding strongly suggests that current T2I models still face fundamental limitations in generating high-precision engineering drawings, as such drawings typically involve strict dimensional accuracy, structural constraints, functional workflows, and complex logical relationships among components. These limitations form the primary bottlenecks for future model improvements. This further reinforces our central claim: in technical image generation scenarios, introducing structured, rubric-based fine-grained evaluation signals is crucial for driving continuous advancements in model capabilities.

\begin{table}[t]
\centering
\caption{\textbf{Statistics of tasks, criteria, and points in TechImage-Bench.} The dataset is divided into three domains, \ie, biology, engineering, and general.}
\setlength{\tabcolsep}{4.5mm}
\begin{tabular}{l|ccc}
\toprule 
 \textbf{Domain} & \textbf{Task}& \textbf{Criteria}& \textbf{Point}\\
\midrule
Biology& $318$& $2772$& $20512$\\
Engineering& $232$& $2427$& $17695$\\
General& $104$&$877$ &$5924$ \\
\midrule
All & $654$ & $6076$ & $44131$\\
\bottomrule
\end{tabular}
\label{tab:dataset}
\end{table}

\begin{table*}[t]
\caption{\textbf{Overall results on TechImage-Bench.} With the exception of Nano Banana 2 in the biology domain, none of the models scored above 0.85 across the three domains of our dataset, indicating that these models still have room for improvement in technical image generation.
}
\label{overall}
\begin{center}
\setlength{\tabcolsep}{4.4mm}
\resizebox{\textwidth}{!}{
\begin{tabular}{lcccccccc}
\toprule
\multicolumn{1}{l}{\multirow{2}{*}{\textbf{Model}}} & \multicolumn{2}{c}{\textbf{Biology}}&\multicolumn{2}{c}{\textbf{Engineering}}&\multicolumn{2}{c}{\textbf{General} }&\multicolumn{2}{c}{\textbf{Overall}}\\
\cmidrule(r){2-3} \cmidrule(r){4-5} \cmidrule(r){6-7} \cmidrule(r){8-9} & {\hspace*{3pt}Acc}&{\hspace*{3pt}Score}& {\hspace*{3pt}Acc}&{\hspace*{3pt}Score}& {\hspace*{3pt}Acc}&{\hspace*{3pt}Score}& {\hspace*{3pt}Acc}&{\hspace*{3pt}Score}\\
\midrule
\makecell[l]{GPT-Image~\cite{GPT-4o}}& $0.704$& $0.425$& $0.556$& $0.258$& $0.718$&$0.463$&$0.660$&$0.382$ \\
\makecell[l]{Nano Banana~\cite{Gemini2_5_Flash}}& $0.697$& $0.400$& $0.579$& $0.276$&$0.716$ &$0.468$&$0.664$&$ 0.381$ \\
\makecell[l]{FLUX~\cite{labs2025flux}}& $0.592$& $0.286$& $0.444$& $0.167$& $0.616$&$0.359$&$0.551$&$0.270$ \\
\makecell[l]{Imagen-3~\cite{baldridge2024imagen}}& $0.600$& $0.288$& $0.492$& $0.195$&$0.638$ &$0.377$&$0.577$&$ 0.287$ \\
\makecell[l]{Seedream~\cite{seedream2025seedream}}& $0.680$& $0.393$&$0.560$&$0.260$&$0.688$&$0.442$&$0.642$&$0.365$ \\
\makecell[l]{Wan2.5~\cite{WAN2_5_T2I_PREVIEW}}& $0.714$& $0.433$& $0.606$& $0.309$&$0.755$ &$0.519$&$0.692$&$ 0.420$ \\
\makecell[l]{GPT-Image-1.5~\cite{GPT-Image-1.5}}& $0.832$& $0.606$& $0.693$& $0.409$& $0.769$&$0.550$&$0.765$&$0.522$ \\
\makecell[l]{Nano Banana Pro~\cite{Nano_Banana_Pro}}& $0.849$& $0.625$& $0.708$& $0.434$&$0.816$ &$0.601$&$0.791$&$ 0.553$ \\
\makecell[l]{Nano Banana 2~\cite{Nano_Banana_2}}& $0.865$& $0.657$& $0.709$& $0.432$&$0.828$ &$0.640$&$0.801$&$ 0.576$ \\
\bottomrule
\end{tabular}
}
\end{center}
\label{tab:overall}
\end{table*}

\begin{figure}[h!]
    \centering
   \vskip -0.1in 
    \includegraphics[width=0.5\textwidth]{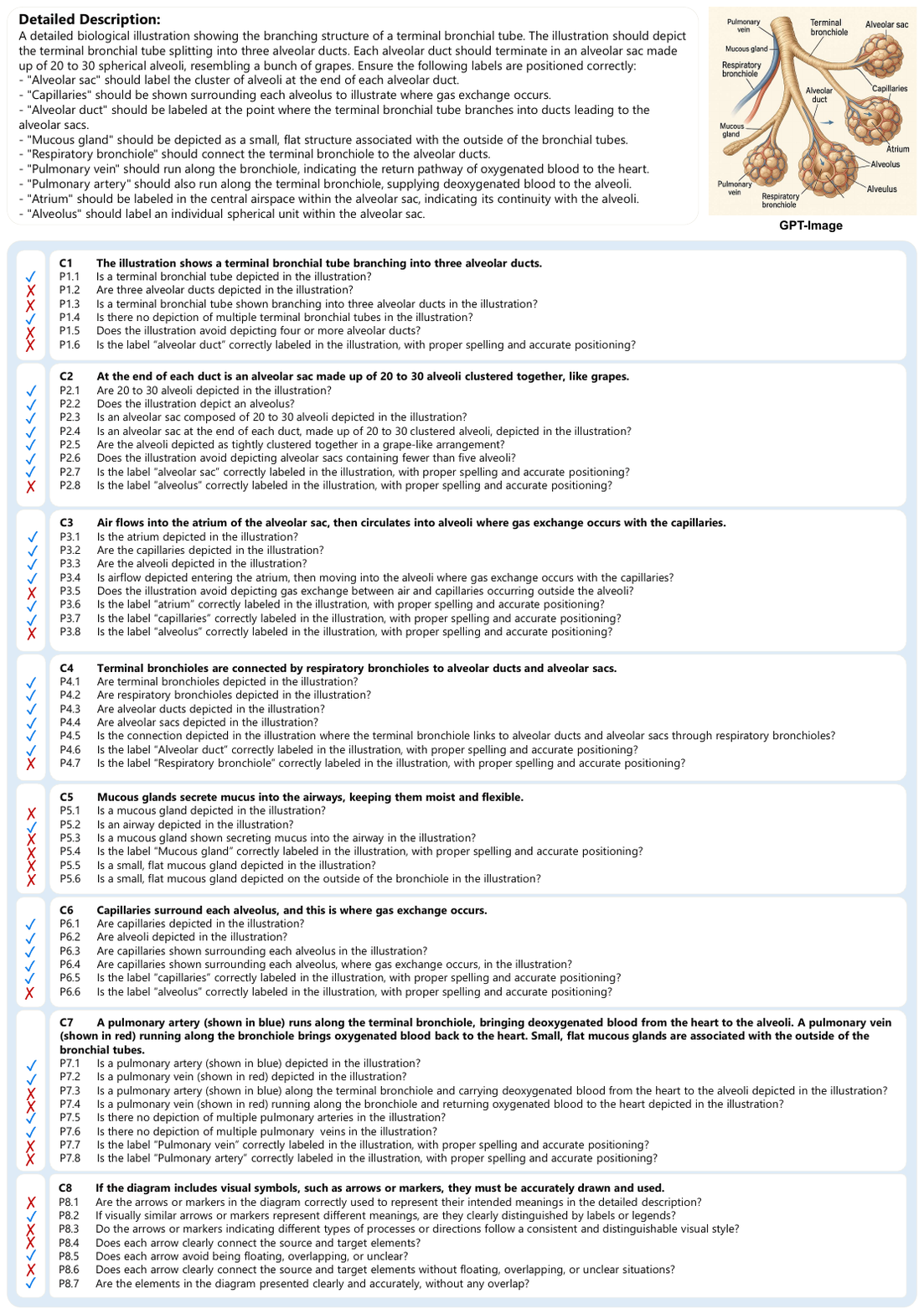}
    \caption{\textbf{Case of evaluation result.} The biological illustration generated by GPT-Image did not depict the three alveolar ducts as required, and failed to represent the mucous glands as small, flat structures on the outside of the bronchial tubes.}
    \label{fig:case_eval}
        \vskip -0.2in 
\end{figure}

\subsection{Qualitative Results}

To more intuitively demonstrate the fine-grained diagnostic capability of TechImage-Bench, Figures \ref{fig:dataset} and \ref{fig:case_eval} present a complete example derived from the dataset and evaluation. This example includes a detailed task description and eight contextual scoring criteria. We can observe that the matrix of satisfied and unsatisfied judgments clearly reveals the fine-grained decision-making process of the LMM evaluator. Instead of providing a single holistic conclusion, the evaluator assesses a series of more granular sub-criteria, enabling a meticulous analysis of generation quality. Compared with approaches that provide only an overall score, TechImage-Bench's multi-level and interpretable scoring scheme delivers richer and more diagnostic feedback, allowing for a more precise characterization of the limitations of current models in technical image generation. For cases of benchmark and evaluation, refer to \textbf{Supplementary Materials}.

\begin{figure*}[hpt!]
    \centering
    \includegraphics[width=1.0\textwidth]{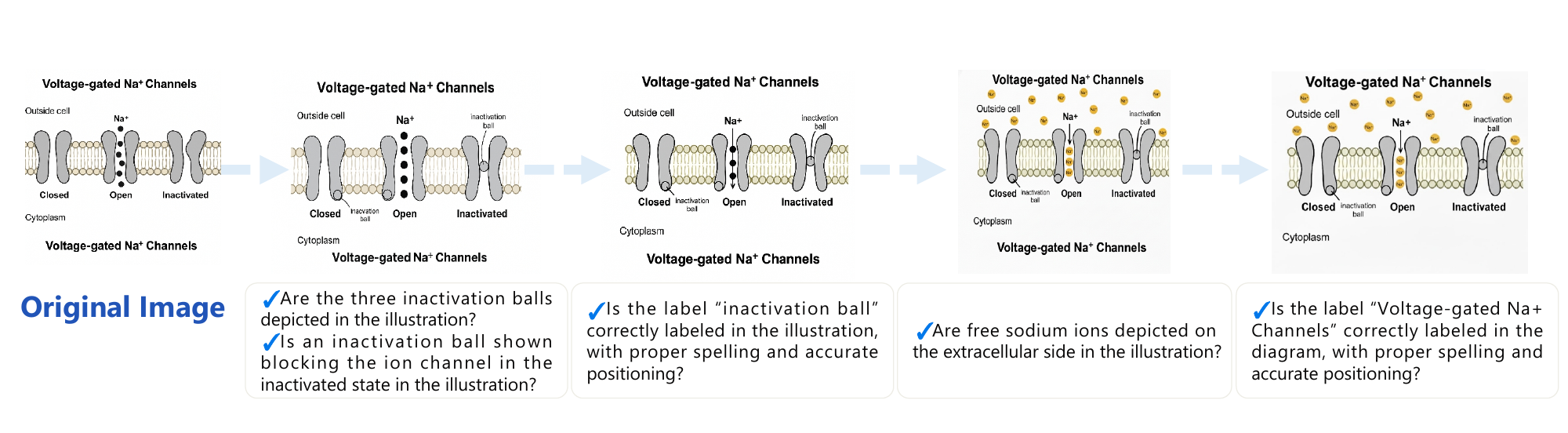}
    \caption{\textbf{Case of iterative refinement.} As the number of editing iterations increases, the generated images gradually meet more grading prompts. Additionally, sometimes a single edit can address multiple grading prompts.}
    \label{fig:case_edit}
\end{figure*}

\subsection{Quality Review of Dataset}

To examine whether our proposed rubrics effectively capture the core criteria that experts rely on when assessing professional quality, we recruited 20 experts with relevant professional backgrounds to systematically evaluate a randomly sampled set of rubrics. Each annotator was asked to rate the quality of each rubric on a 1–5 scale, ranging from 5 for completely error-free and unambiguous to 1 for factually incorrect or contradictory to the original text. A score of 4 indicates no obvious errors or ambiguities and represents the minimum acceptable quality level we set. The results in Table \ref{tab:human} demonstrate the high quality and professionalism of our TechImage-Bench.

\begin{table}[t]
\centering
\caption{\textbf{Human rating results and satisfactory ratio.} This proves the high quality of our benchmark.
}
\setlength{\tabcolsep}{5.0mm}
\begin{tabular}{l|cc}
\toprule
 \textbf{Domain} & \textbf{Rating} & \textbf{Satisfactory Ratio} \\
\midrule
Biology & $4.0986$& $82.00\%$ \\
Engineering & $4.0964$ & $84.21\%$ \\
General & $4.1045$ & $84.85\%$ \\
\bottomrule
\end{tabular}
\label{tab:human}
\end{table}


\subsection{Consistency of LMM Evaluator}

Verifying the consistency between LMM evaluation results and human judgments is crucial. To this end, we conducted an additional human annotation study on the experimental results and compared the resulting scores with the evaluations produced by the LMM. We then calculated the score differences across multiple percentiles. Table \ref{tab:consis} presents the range of discrepancies between human and LMM-based evaluations. The results show a high level of agreement between the LMM evaluator and human judgments. This indicates that even for the most ambiguous or complex samples, the LMM's evaluation results remain stable and highly consistent with human reviewers, without noticeable bias, thereby providing strong support for the objectivity and reliability of this benchmark evaluation.

\begin{table}[t]
\centering
\caption{\textbf{Consistency of LMM evaluator.} The model's scores are very close to human evaluation results, suggesting that our evaluation results align closely with human judgment.}
\setlength{\tabcolsep}{3.4mm}
\begin{tabular}{c|cc}
\toprule
\textbf{Percentile} & \textbf{Acc Difference} & \textbf{Score Difference} \\
\midrule
$50\%$ & $0.0096$ & $0.0046$ \\
$80\%$ & $0.0400$ & $0.0312$ \\
$90\%$ & $0.0528$ & $0.0422$ \\
$100\%$ & $0.1000$ & $0.0625$ \\
\bottomrule
\end{tabular}
\label{tab:consis}
\end{table}

\begin{table*}[t]
\caption{\textbf{Refinement results with rubrics.} As the number of editing iterations increases, the scores across all three domains gradually improve, demonstrating that rubrics can be used as explicit and executable guidance. }
    \vskip -0.2in 
\label{edit_overall}
\begin{center}
\setlength{\tabcolsep}{5.1mm}
\resizebox{\textwidth}{!}{
\begin{tabular}{lcccccccc}
\toprule
\multicolumn{1}{l}{\multirow{2}{*}{\textbf{Number}}} & \multicolumn{2}{c}{\textbf{Biology}}&\multicolumn{2}{c}{\textbf{Engineering}}&\multicolumn{2}{c}{\textbf{General} }&\multicolumn{2}{c}{\textbf{Overall}}\\
\cmidrule(r){2-3} \cmidrule(r){4-5} \cmidrule(r){6-7} \cmidrule(r){8-9} 
& {\hspace*{3pt}Acc}&{\hspace*{3pt}Score}& {\hspace*{3pt}Acc}&{\hspace*{3pt}Score}& {\hspace*{3pt}Acc}&{\hspace*{3pt}Score}& {\hspace*{3pt}Acc}&{\hspace*{3pt}Score}\\
\midrule
\makecell[l]{Original}     & $0.704$ & $0.425$ & $0.556$ & $0.258$ & $0.718$ & $0.463$ & $0.660$ & $0.382$ \\
\midrule
\makecell[l]{N=1}          & $0.783$ & $0.518$ & $0.639$ & $0.340$ & $0.765$ & $0.585$ & $0.729$ & $0.481$ \\
\makecell[l]{N=2}          & $0.820$ & $0.579$ & $0.687$ & $0.397$ & $0.801$ & $0.629$ & $0.769$ & $0.535$ \\
\makecell[l]{N=3}          & $0.847$ & $0.629$ & $0.723$ & $0.442$ & $0.831$ & $0.673$ & $0.800$ & $0.581$ \\
\makecell[l]{N=4}          & $0.863$ & $0.658$ & $0.750$ & $0.480$ & $0.845$ & $0.696$ & $0.820$ & $0.611$ \\
\makecell[l]{N=5}          & $0.874$ & $0.675$ & $0.763$ & $0.504$ & $0.857$ & $0.717$ & $0.831$ & $0.632$ \\
\makecell[l]{N=10}         & $0.903$ & $0.735$ & $0.812$ & $0.597$ & $0.879$ & $0.759$ & $0.865$ & $0.697$ \\
\bottomrule
\end{tabular}
}
\end{center}
\label{tab:edit}
\end{table*}

\section{Discussion}

\subsection{Refinement with Rubrics}

Rubrics can serve not only as tools for diagnosing fine-grained model errors but also as effective and actionable supervision signals for improving image quality. We conducted an iterative image refinement experiment based on rubrics. In each iteration, we extracted all points with a score of 0 from the rubrics and used these failure cases as explicit editing instructions fed back to the editing model to modify the image generated in the previous round. We then recorded the best results produced during the editing process. Specifically, in each iteration, for every criterion with a score of 0, we used those points together with the original image to generate editing instructions via an LMM, directing Nano Banana to perform the edits.

\subsection{Overall Refinement Results}

Using the original image generated by GPT-Image as the baseline, we conducted the experiment. As shown in Table \ref{tab:edit}, after just one iteration, the accuracy and score increased markedly, indicating that accurately identifying key errors can significantly improve image quality. As the editing proceeded, the overall quality continued to increase substantially. This demonstrates that when rubrics are used as explicit and executable guidance signals, they can continually enhance fine-grained correctness across tasks.

\subsection{Case Study of Refinement}

As shown in Figure \ref{fig:case_edit}, using our editing process, the model first added an inactivation ball to each ion channel, then introduced an arrow pointing from the extracellular side to the intracellular side, subsequently placed some free sodium ions on the outside of the cell, and finally removed the duplicate labels from the illustration. We observe that (1) as the number of iterations increases, the image progressively addresses more grading prompts, demonstrating that our rubrics effectively improve the quality of the generated illustrations, and (2) during refinement, multiple grading prompts may be simultaneously improved, suggesting the potential of our method.

\section{Limitations}
\label{sec:lim}

Although TechImage-Bench provides a scalable, fine-grained, and highly interpretable evaluation framework for technical image generation, its current domain coverage mainly focuses on biology, engineering, and general scenarios. Expanding the benchmark to additional professional domains in future work will further enhance its breadth and diversity. In addition, while most steps of the rubric construction process are automated, a certain degree of manual filtering and text refinement is still required. Although our experiments show high alignment between rubric quality and human experts, we plan to incorporate external knowledge sources and more advanced reasoning methods to achieve a fully automated benchmark construction pipeline.

\section{Broader Impact}
\label{sec:impact}

TechImage-Bench aims to advance technical image generation toward models with high accuracy and structured understanding capabilities. This direction has potential positive impact across education, publishing, industrial documentation, and patent drafting in reducing repetitive diagram creation, improving productivity, and lowering the barriers to professional visual communication.

\section{Conclusion}

In this paper, we introduced TechImage-Bench, a new benchmark designed to address a critical gap in evaluating text-to-image models for technical image generation. We observed that while current models produced visually impressive results, they frequently exhibited fine-grained, fact-based errors that made them unsuitable for use in high-stakes domains such as science, engineering, and medicine. Standard evaluation metrics failed to capture these crucial inaccuracies. Our core contribution was a decompositional evaluation methodology, which replaced coarse and subjective judgments with a detailed, objective, binary yes-or-no set of evaluation rubrics using an LMM evaluator. Our experiments yielded two key findings. First, comprehensive testing on TechImage-Bench revealed that even state-of-the-art models performed poorly in accurately reproducing fine-grained scientific details. Second, we validated that the evaluation rubrics could serve as actionable feedback for image optimization, guiding models to revise their outputs and significantly improve their scores, revealing the potential of AI for precise and technical visual generation. For ethics statement, refer to \textbf{Supplementary Materials}. 

{
    \small
    \bibliographystyle{ieeenat_fullname}
    \bibliography{main}
}

\clearpage
\setcounter{figure}{0}
\setcounter{table}{0}
\setcounter{section}{0}

\renewcommand\thesection{\Roman{section}}
\renewcommand\thefigure{\Alph{figure}}
\renewcommand\thetable{\Alph{table}}

\twocolumn[{
\begin{center}
\includegraphics[width=\textwidth]{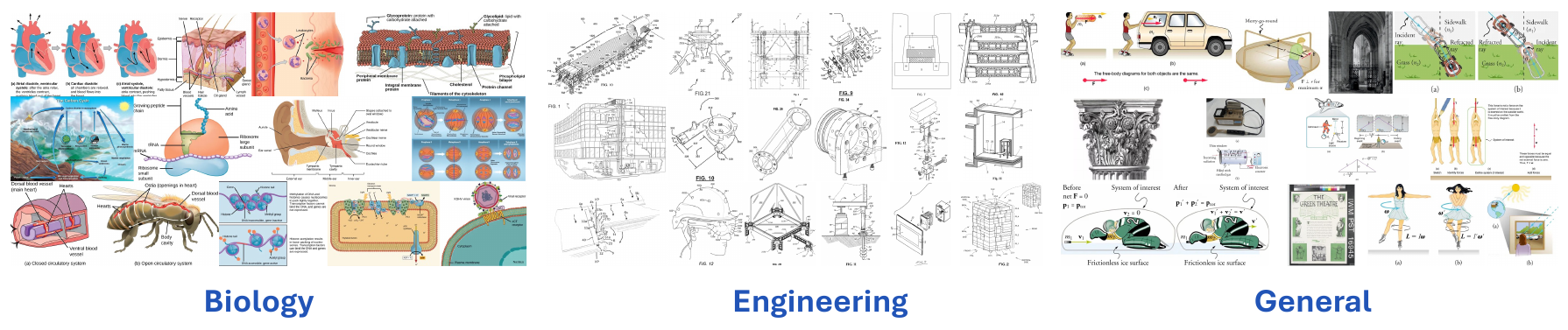}
\captionof{figure}{Overview of three domains in ProImage-Bench.}
\label{fig:overview}
\end{center}
}]

This supplemental material mainly contains:
\begin{itemize}
\item Additional details of implementation in Section \ref{sec:more_impl}
\item Additional details of experimental setup in Section \ref{sec:more_eval}
\item Additional details of dataset  in Section \ref{sec:more_data}
\item Extra cases of benchmark in Section \ref{sec:more_case_gen}
\item Extra cases of evaluation in Section \ref{sec:more_case_eval}
\item Online benchmark and code release in Section \ref{sec:release}
\item Ethical statement in Section \ref{sec:ethical}
\end{itemize}
\section{Additional Details of Implementation}
\label{sec:more_impl}

For points expanding, we decompose constraints, \textit{i.e.}, criteria into a set of binary scoring questions that can be explicitly answered with \textit{yes} or \textit{no}. The model is asked to generate a set of binary scoring questions that can be explicitly answered with binary answers.

To further generate grading prompts of rubrics that characterize incorrect drawing behaviors which should not appear in the diagram, we formulate this step as an error-focused prompt generation task. For each label extracted during the label extraction process, we prompt the large model to identify the explicit or implicit constraint most relevant to that label.

All prompts can be found in our code.

\section{Additional Details of Experimental Setup}
\label{sec:more_eval}

\paragraph{Image Generation}

For image generation, we provide the following prompt to the image generation models:
\texttt{Generate a professional-style image strictly following the provided detailed description, without modifying any of the instructions. Detailed description: \{detailed description\} Ensure the perspective view allows clear visibility of all labeled components.}

\paragraph{Human Quality Evaluation}
In the quality evaluation of the dataset, we recruited 20 experts with relevant domain knowledge to review the rubrics. Annotators were asked to rate each rubric on a 1–5 scale. A score of 5 indicates complete consistency with no ambiguity. A score of 4 indicates no apparent errors. A score of 3 reflects minor inconsistencies, though the intended meaning can still be inferred using the full diagram and background knowledge. A score of 2 indicates the presence of errors or contradictions that are difficult to resolve even with background knowledge, but do not involve factual mistakes or direct conflicts with the original text. A score of 1 denotes clear factual errors or direct contradictions with the original text.

\section{Additional Details of Dataset}
Our dataset is divided into three domains, as shown in Figure \ref{fig:overview}. The biology domain contains illustrations sourced from biology textbooks, covering a wide range of scientific content, including cellular and organelle structures, physiological and metabolic processes, and ecological or evolutionary diagrams. The engineering domain consists of illustrations derived from patent drawings, depicting various architectural and mechanical structures. The general domain primarily includes schematic diagrams, as well as poster and architectural images.
\label{sec:more_data}

\section{Extra Cases of Benchmark}
\label{sec:more_case_gen}

\begin{figure*}[h!]
    \centering
    \includegraphics[width=0.9\textwidth]{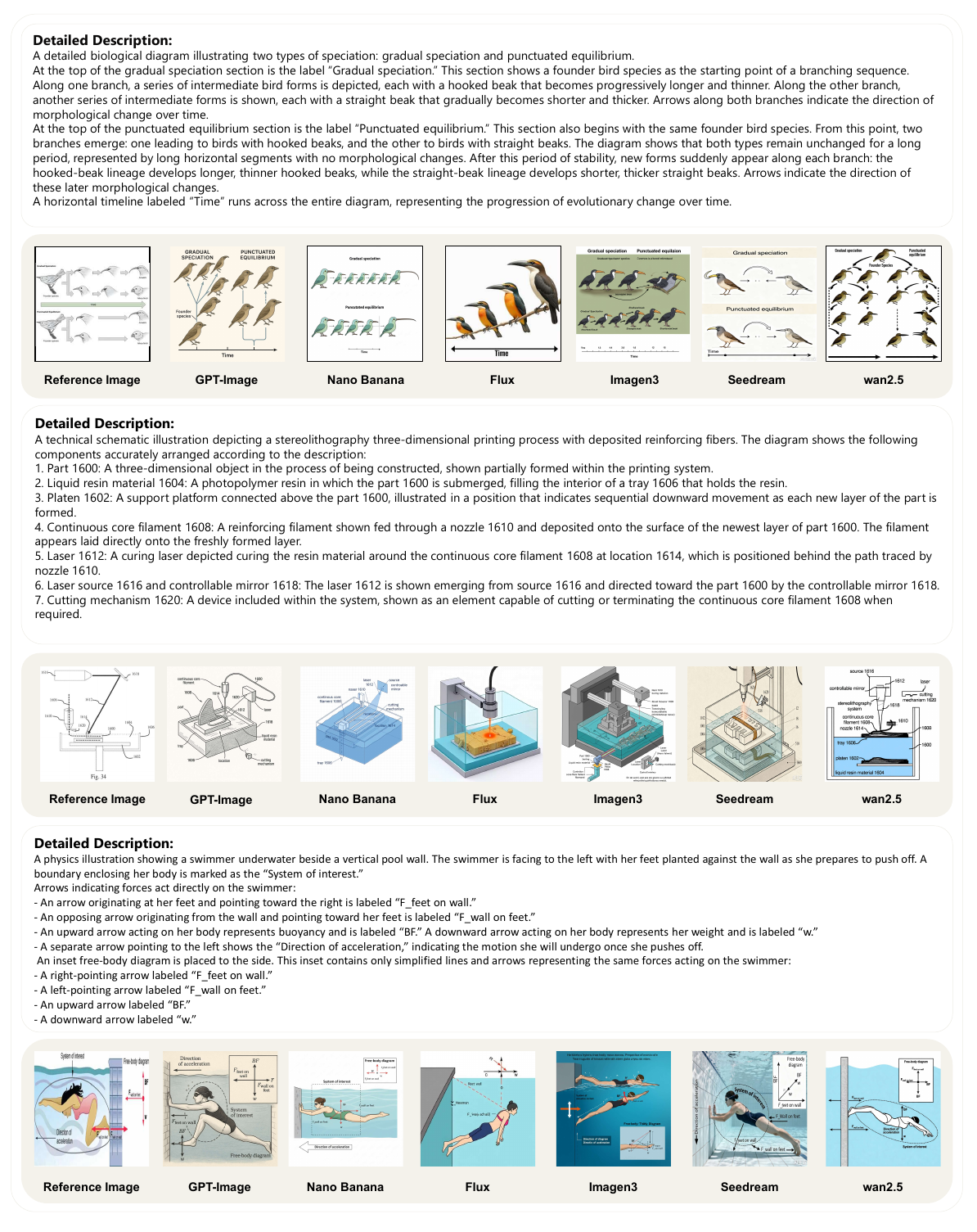}
    \caption{Generation cases of benchmark (Group 1).}
    \label{fig:G1}
\end{figure*}

\begin{figure*}[h!]
    \centering
    \includegraphics[width=0.9\textwidth]{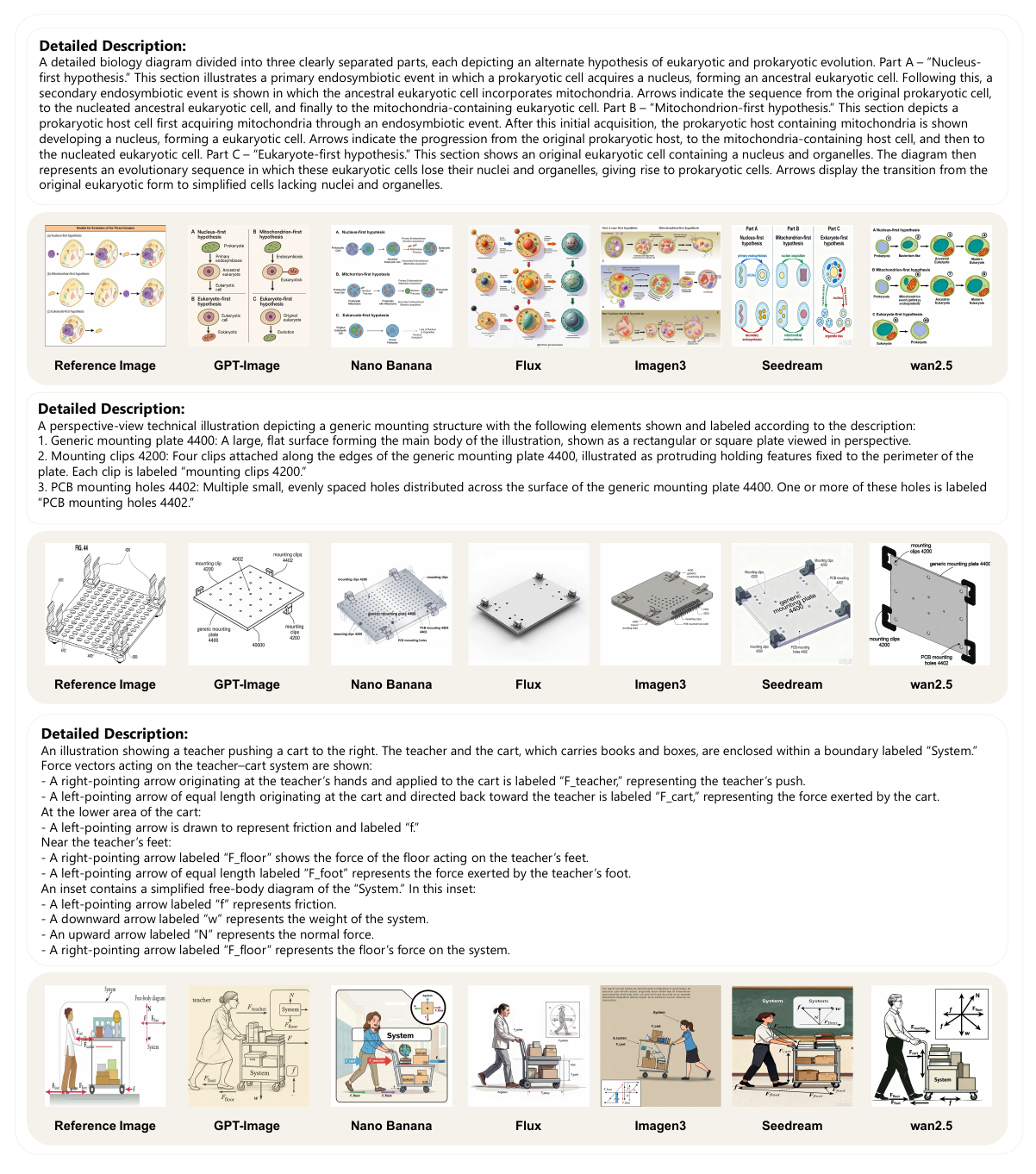}
    \caption{Generation cases of benchmark (Group 2).}
    \label{fig:G2}
\end{figure*}

\begin{figure*}[h!]
    \centering
    \includegraphics[width=0.9\textwidth]{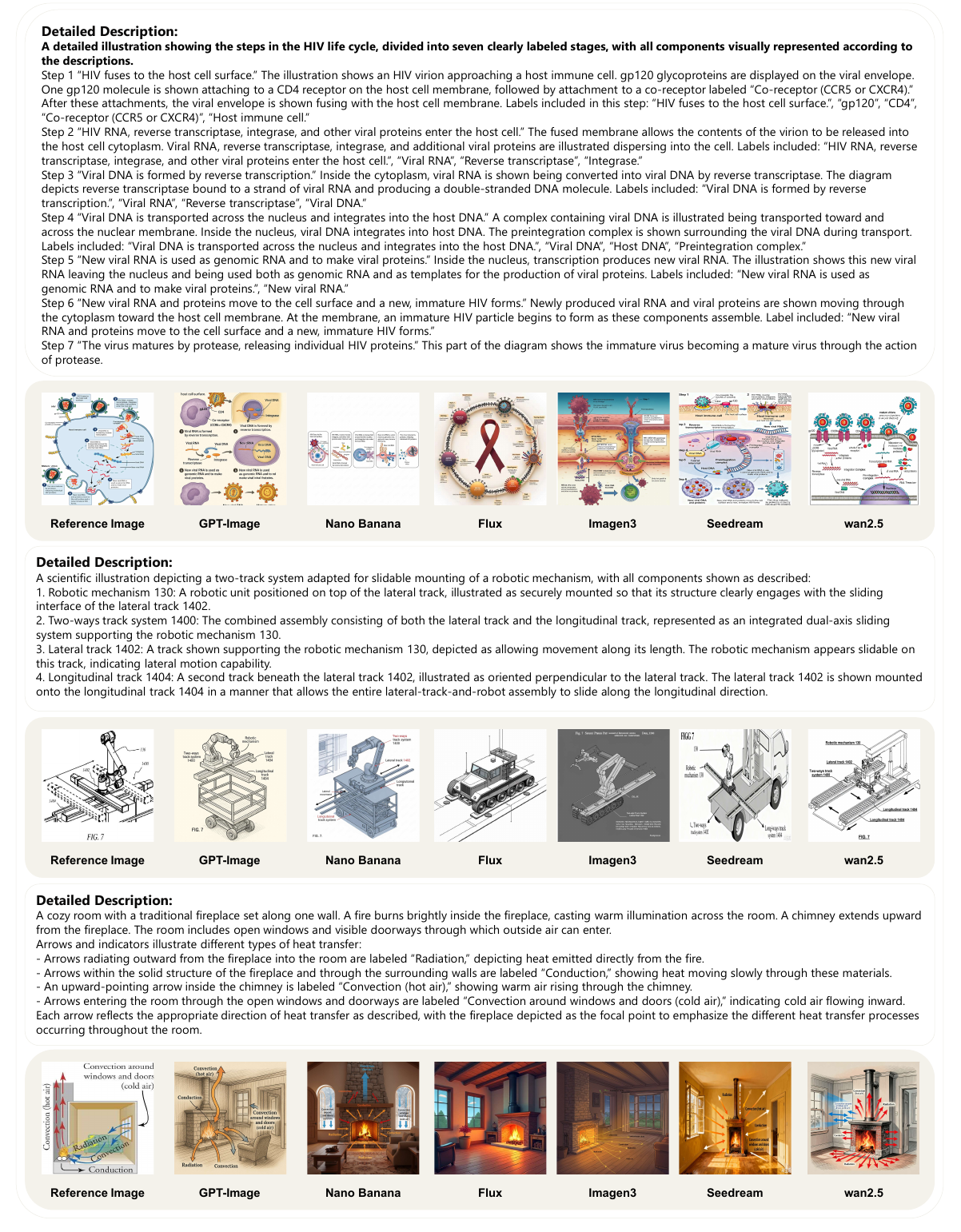}
    \caption{Generation cases of benchmark (Group 3).}
    \label{fig:G3}
\end{figure*}

\begin{figure*}[h!]
    \centering
    \includegraphics[width=0.9\textwidth]{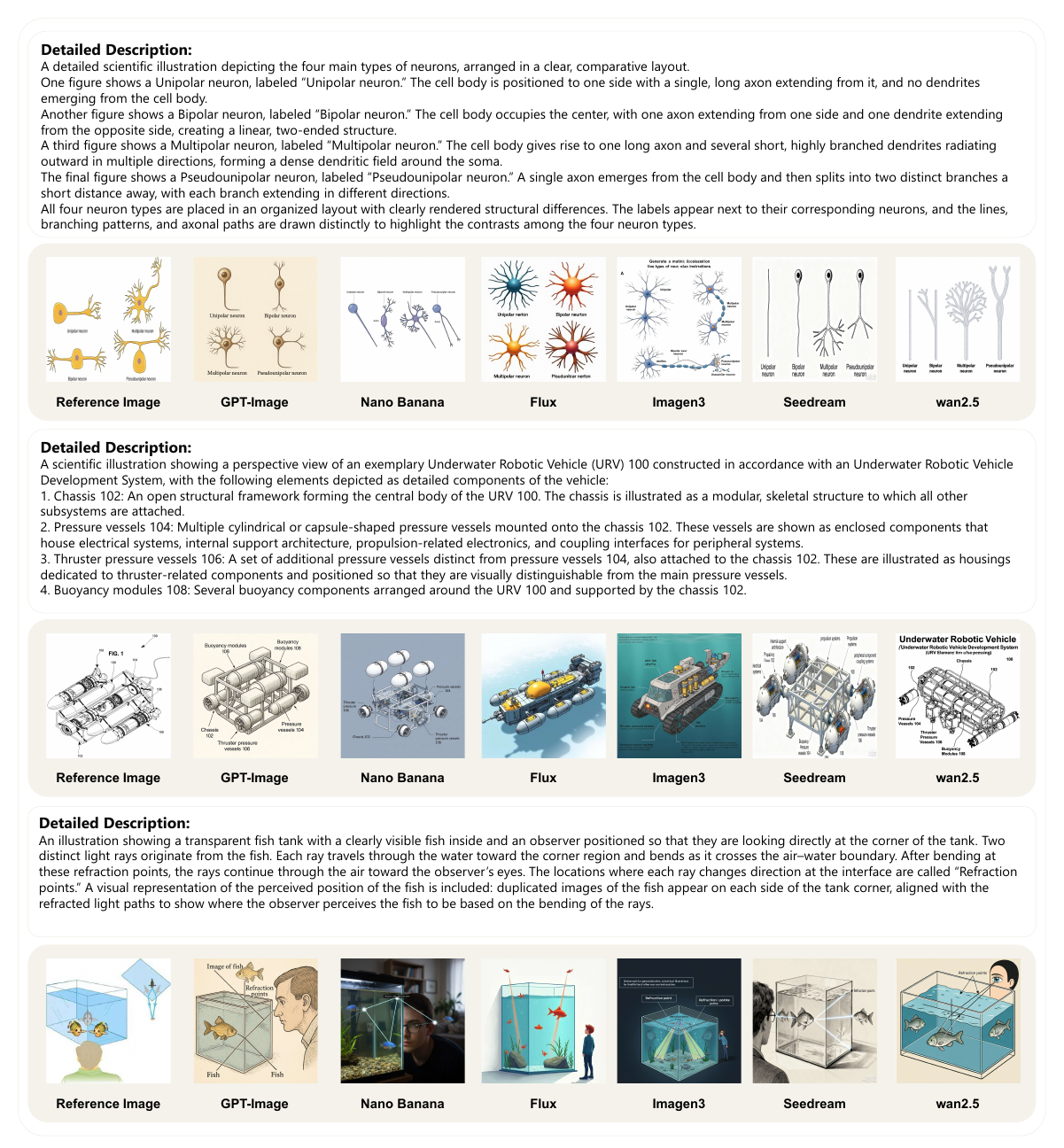}
    \caption{Generation cases of benchmark (Group 4).}
    \label{fig:G4}
\end{figure*}

\begin{figure*}[h!]
    \centering
    \includegraphics[width=0.9\textwidth]{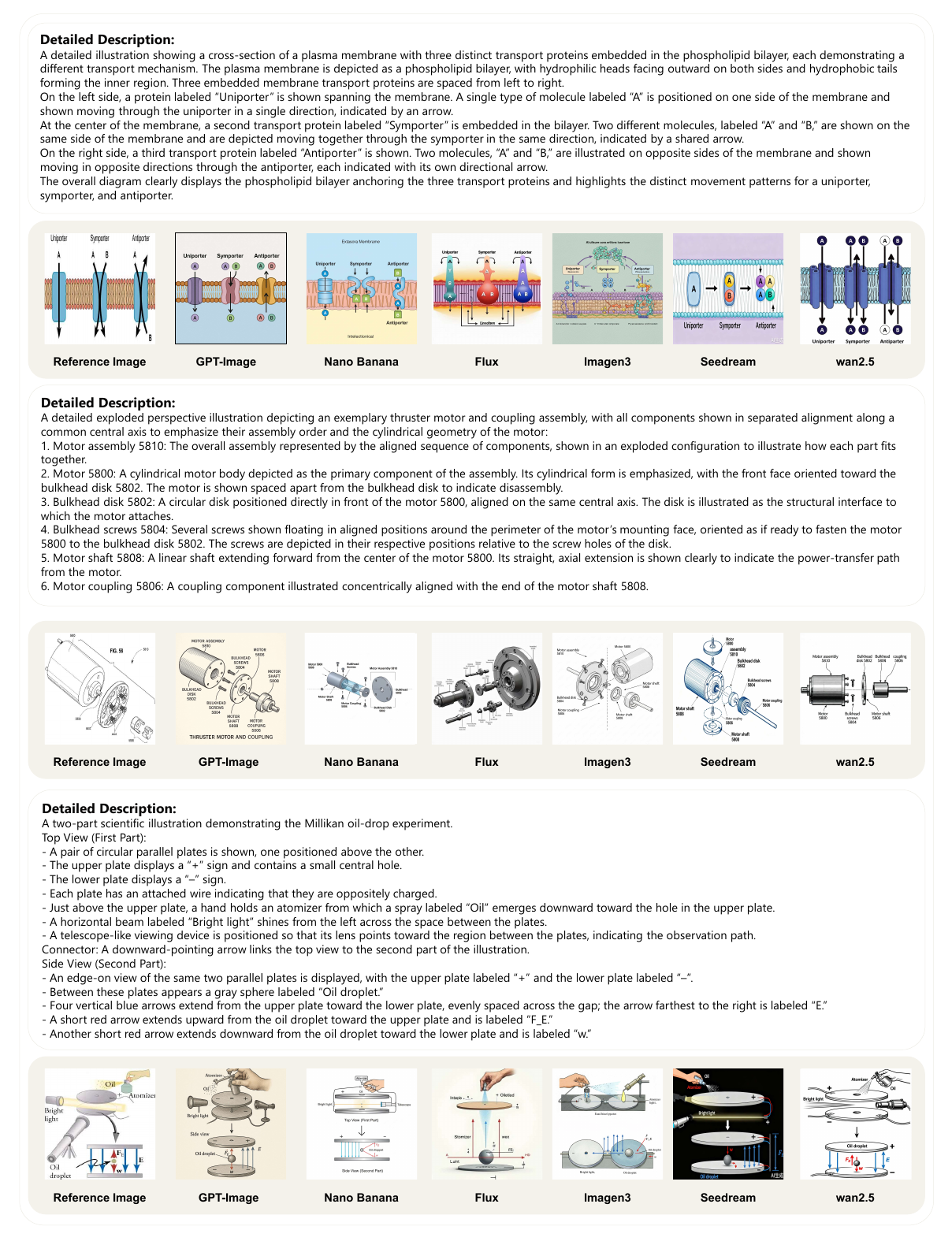}
    \caption{Generation cases of benchmark (Group 5).}
    \label{fig:G5}
\end{figure*}

\begin{figure*}[h!]
    \centering
    \includegraphics[width=0.9\textwidth]{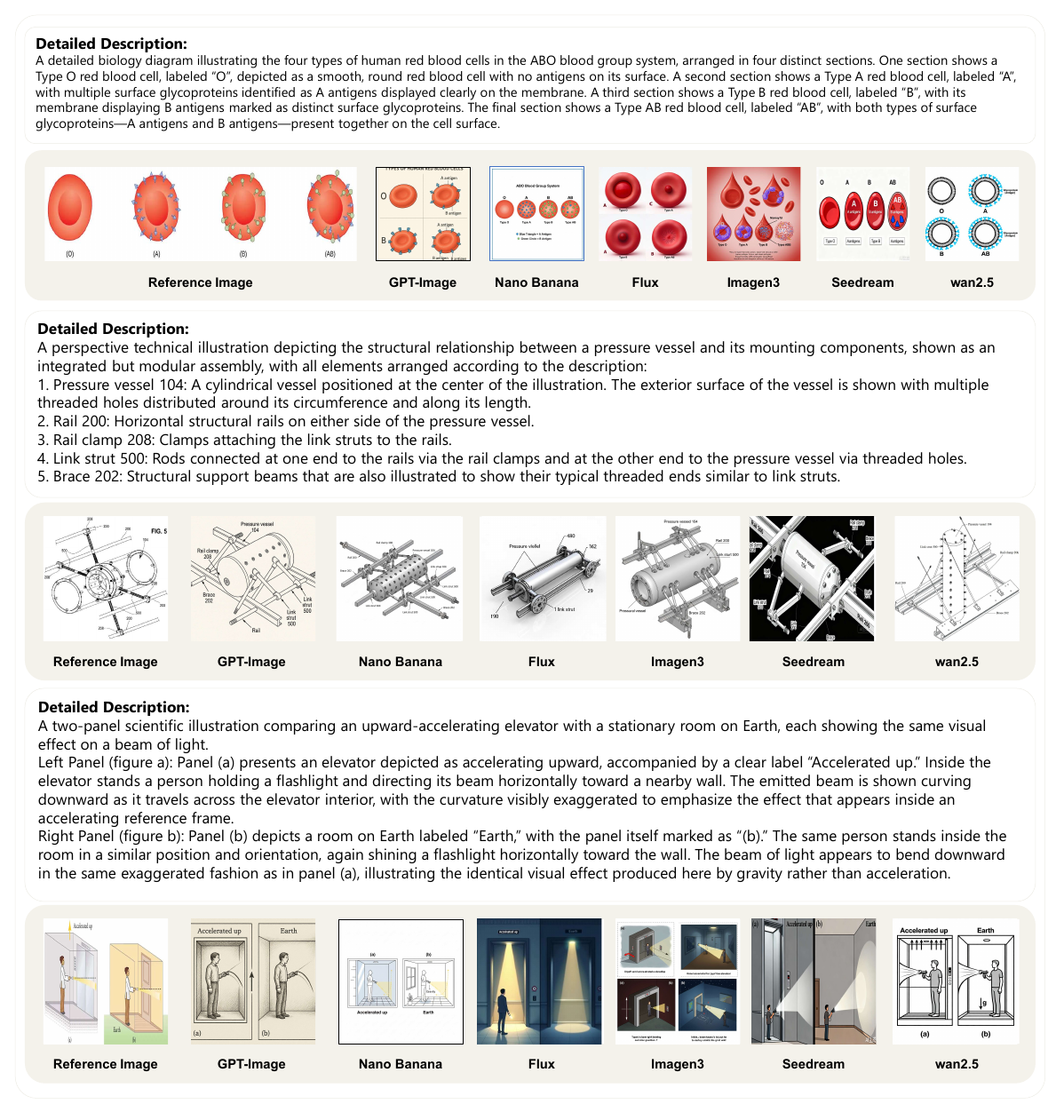}
    \caption{Generation cases of benchmark (Group 6).}
    \label{fig:G6}
\end{figure*}

\begin{figure*}[h!]
    \centering
    \includegraphics[width=0.9\textwidth]{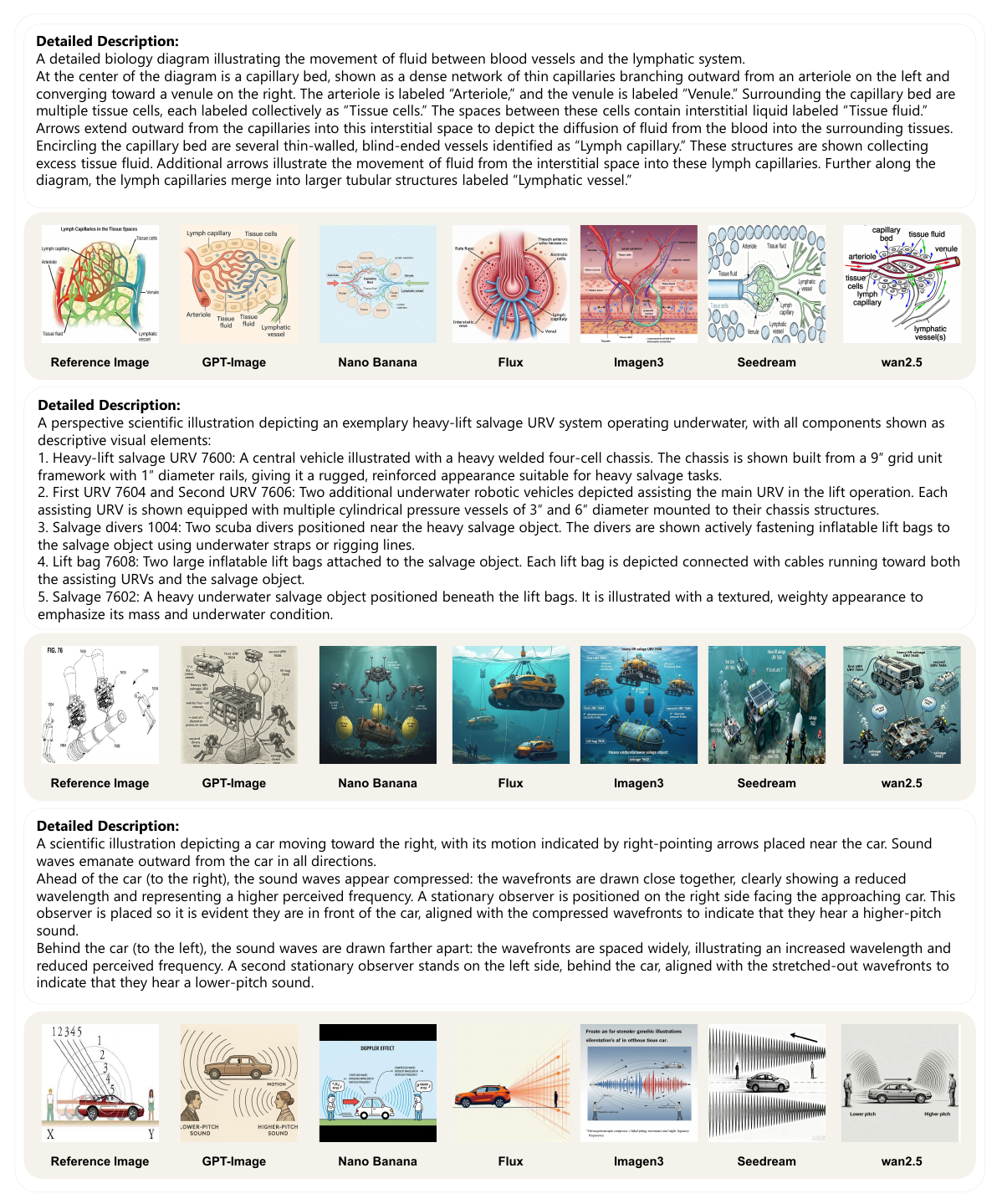}
    \caption{Generation cases of benchmark (Group 7).}
    \label{fig:G7}
\end{figure*}

\begin{figure*}[h!]
    \centering
    \includegraphics[width=0.9\textwidth]{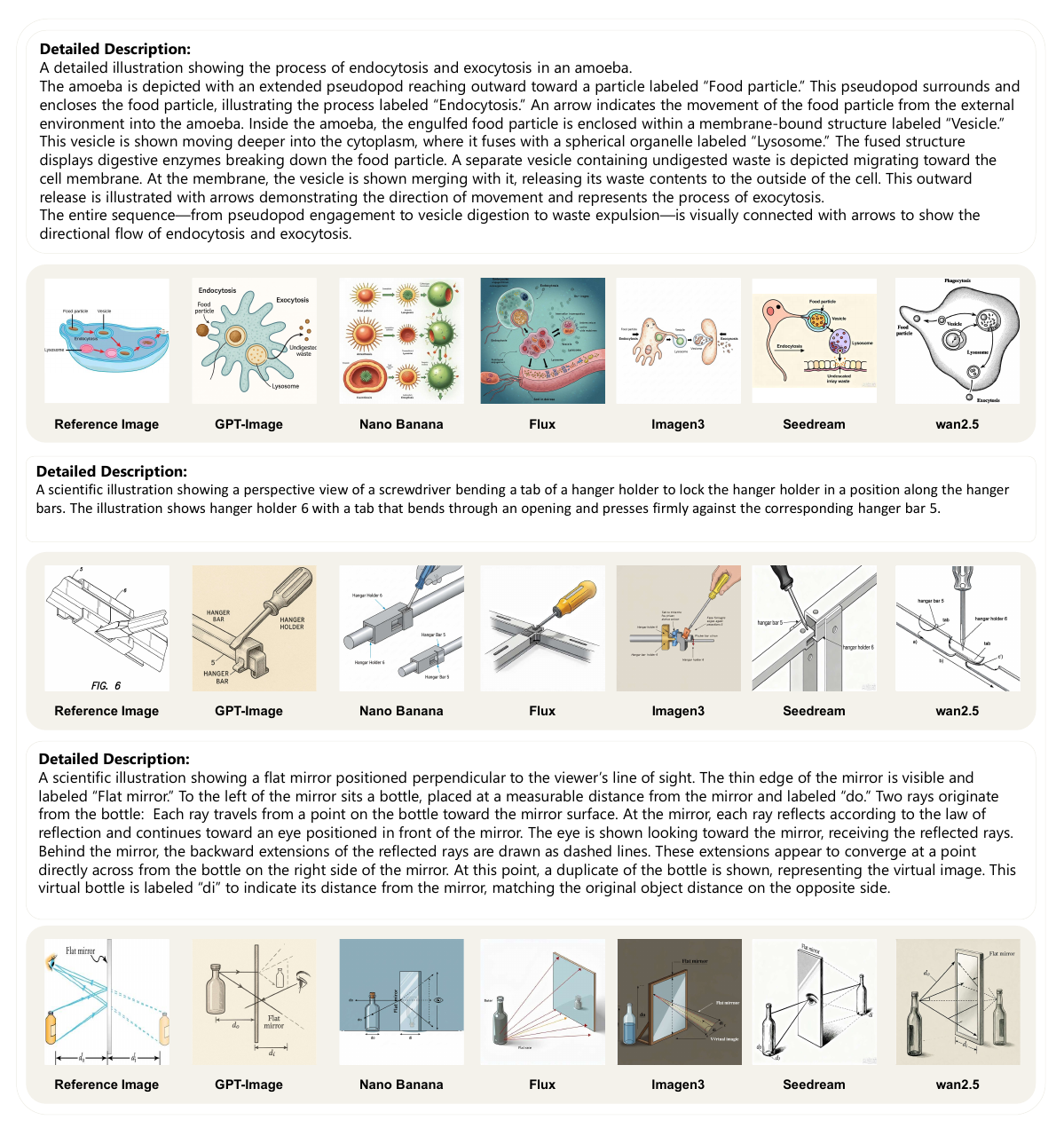}
    \caption{Generation cases of benchmark (Group 8).}
    \label{fig:G8}
\end{figure*}

\begin{figure*}[h!]
    \centering
    \includegraphics[width=0.9\textwidth]{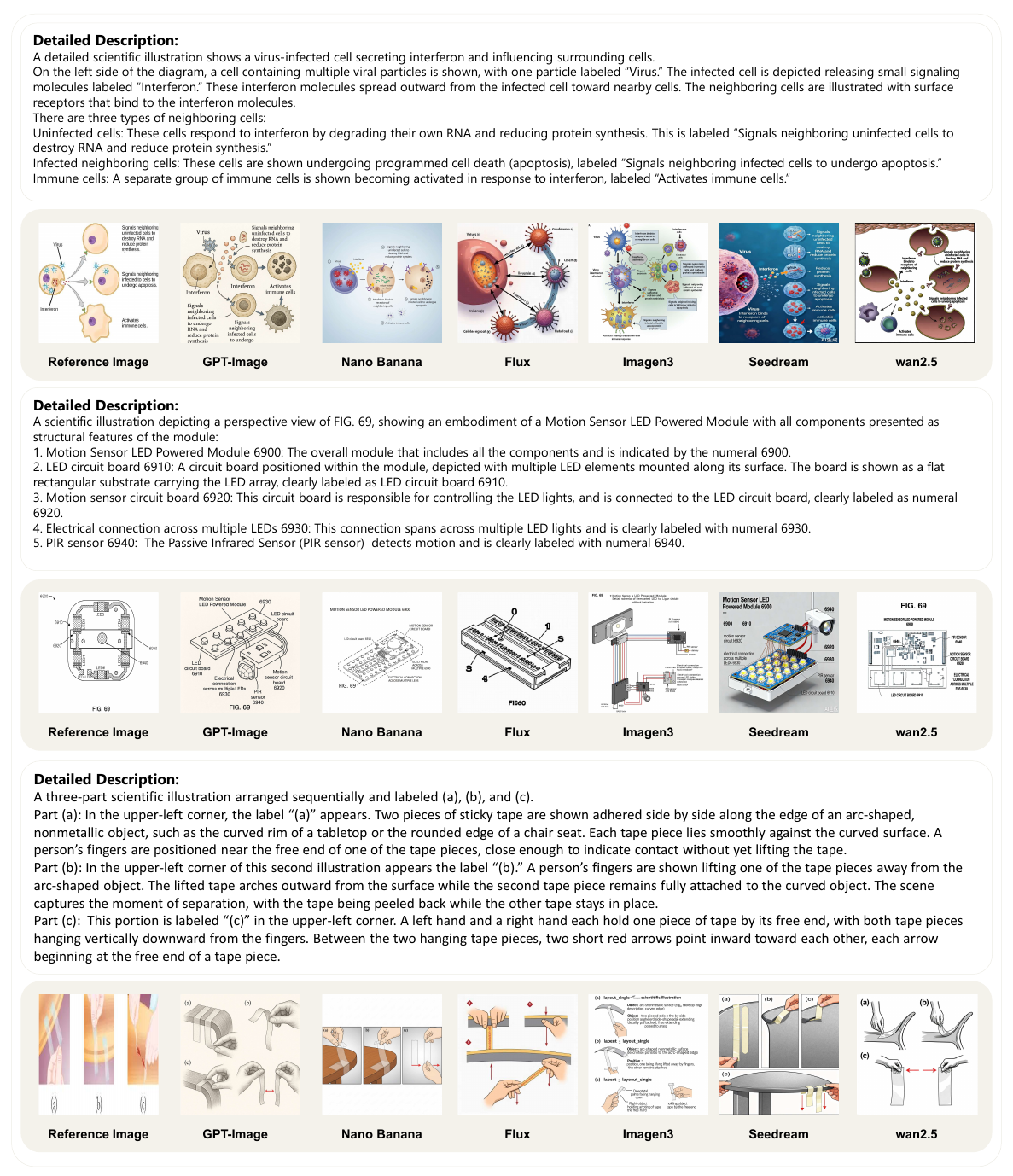}
    \caption{Generation cases of benchmark (Group 9).}
    \label{fig:G9}
\end{figure*}

\begin{figure*}[h!]
    \centering
    \includegraphics[width=0.9\textwidth]{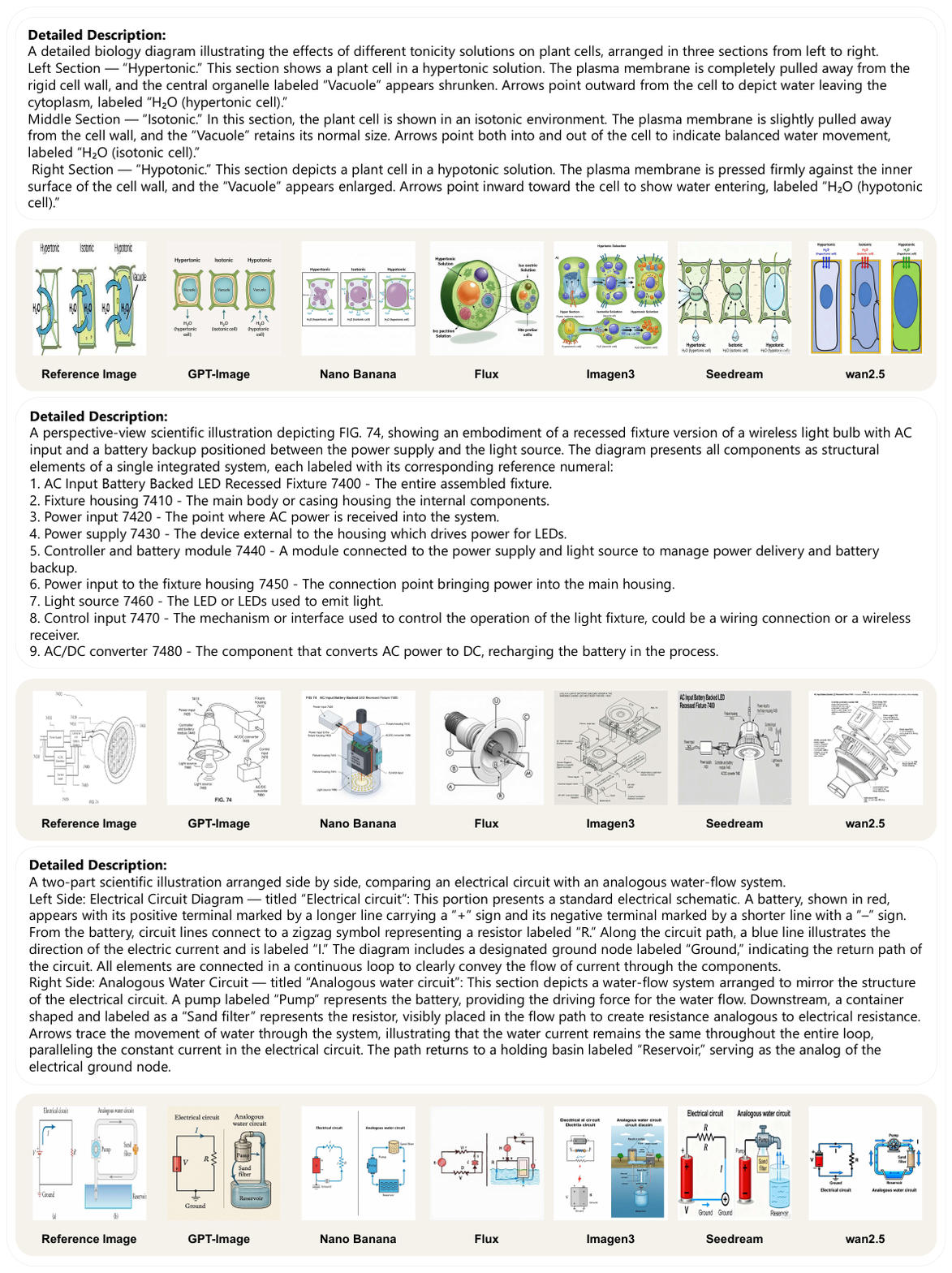}
    \caption{Generation cases of benchmark (Group 10).}
    \label{fig:G10}
\end{figure*}

We provide extensive examples of current model outputs on the benchmark. Please refer to Figures \ref{fig:G1}-\ref{fig:G10}.

\section{Extra Cases of Evaluation}
\label{sec:more_case_eval}

\begin{figure*}[h!]
    \centering
    \includegraphics[width=0.85\textwidth]{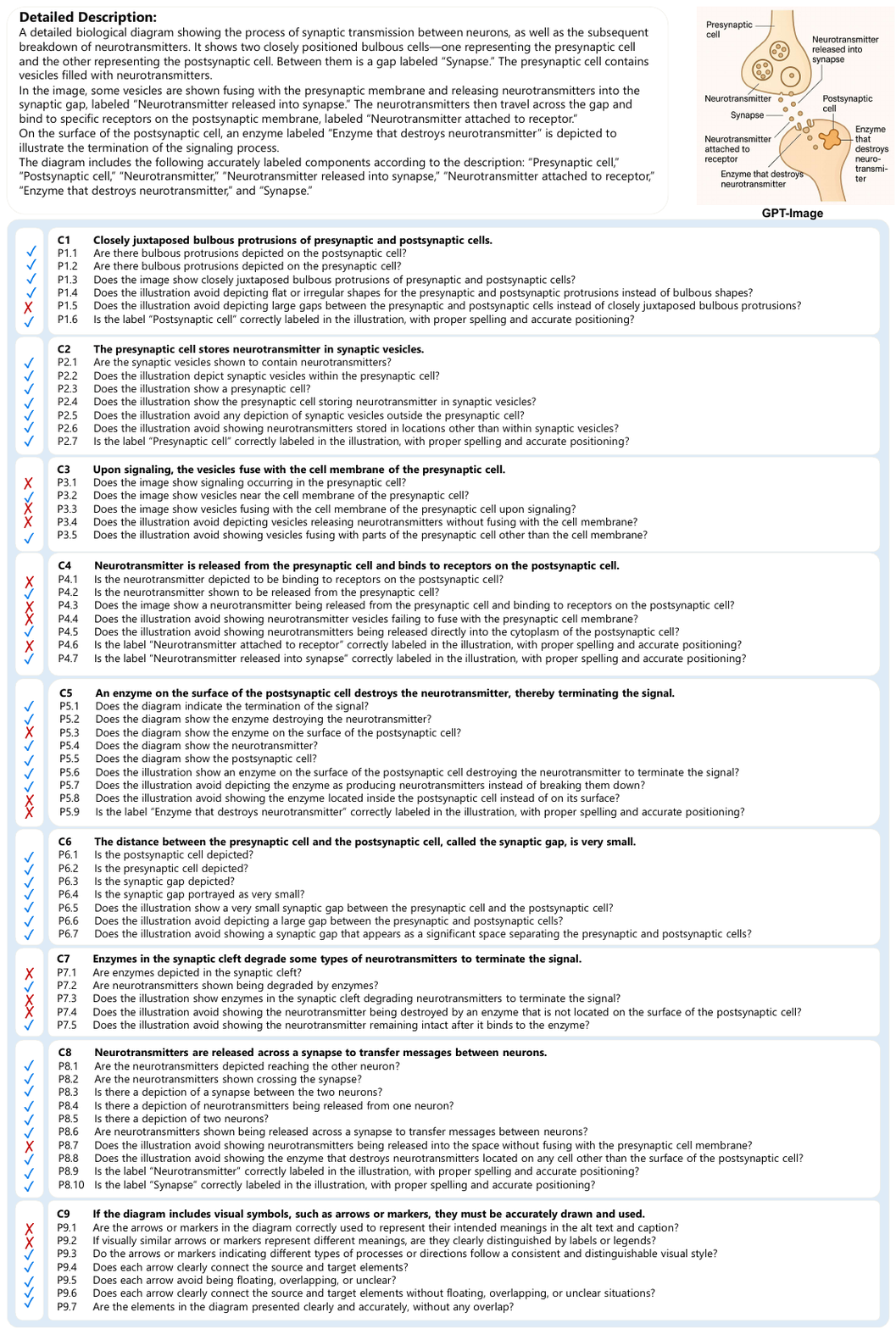}
    \caption{Evaluation case of GPT-4o in the first task.}
    \label{fig:E1-1}
\end{figure*}

\begin{figure*}[h!]
    \centering
    \includegraphics[width=0.85\textwidth]{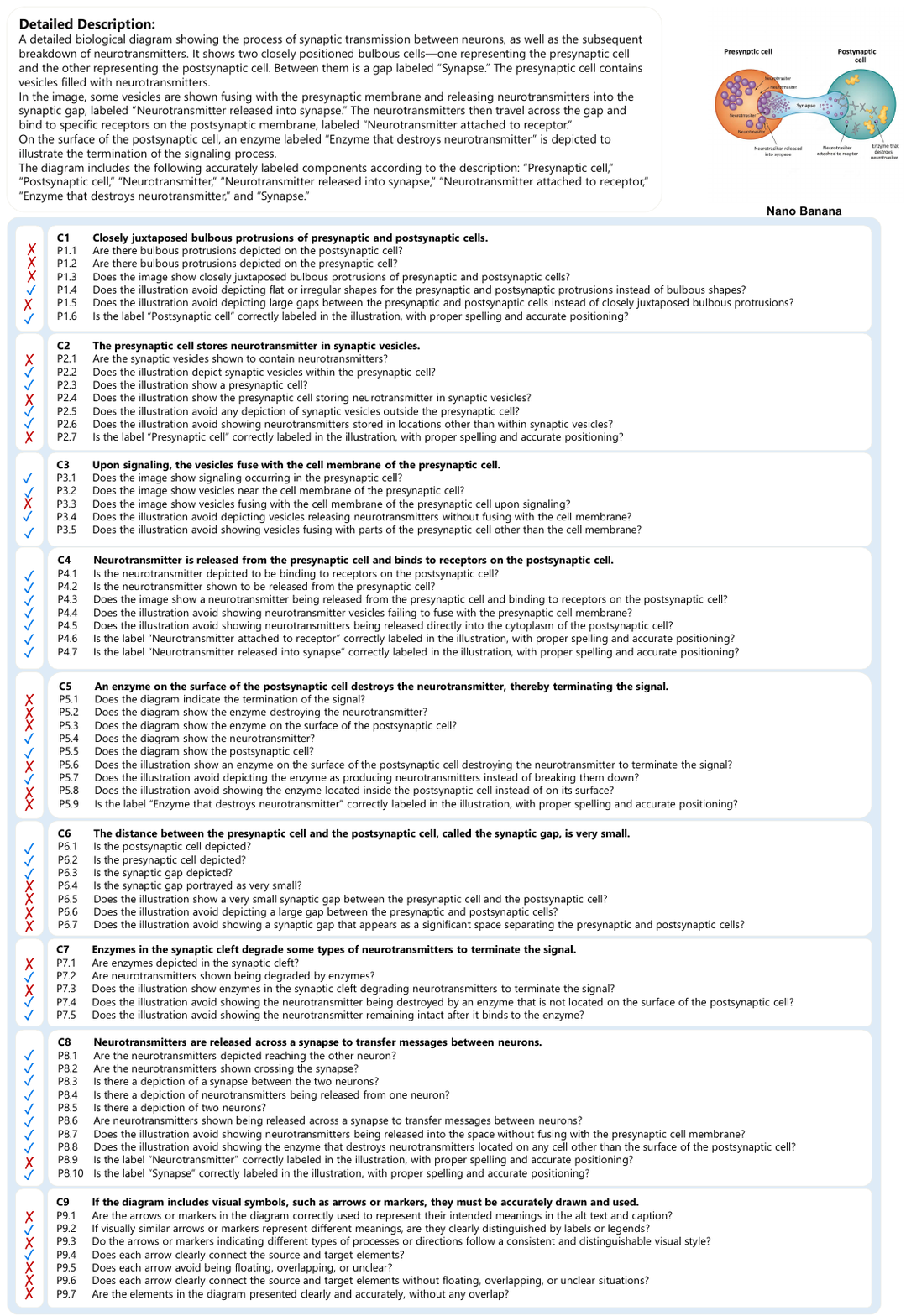}
    \caption{Evaluation case of Nano Banana in the first task.}
    \label{fig:E1-2}
\end{figure*}

\begin{figure*}[h!]
    \centering
    \includegraphics[width=0.85\textwidth]{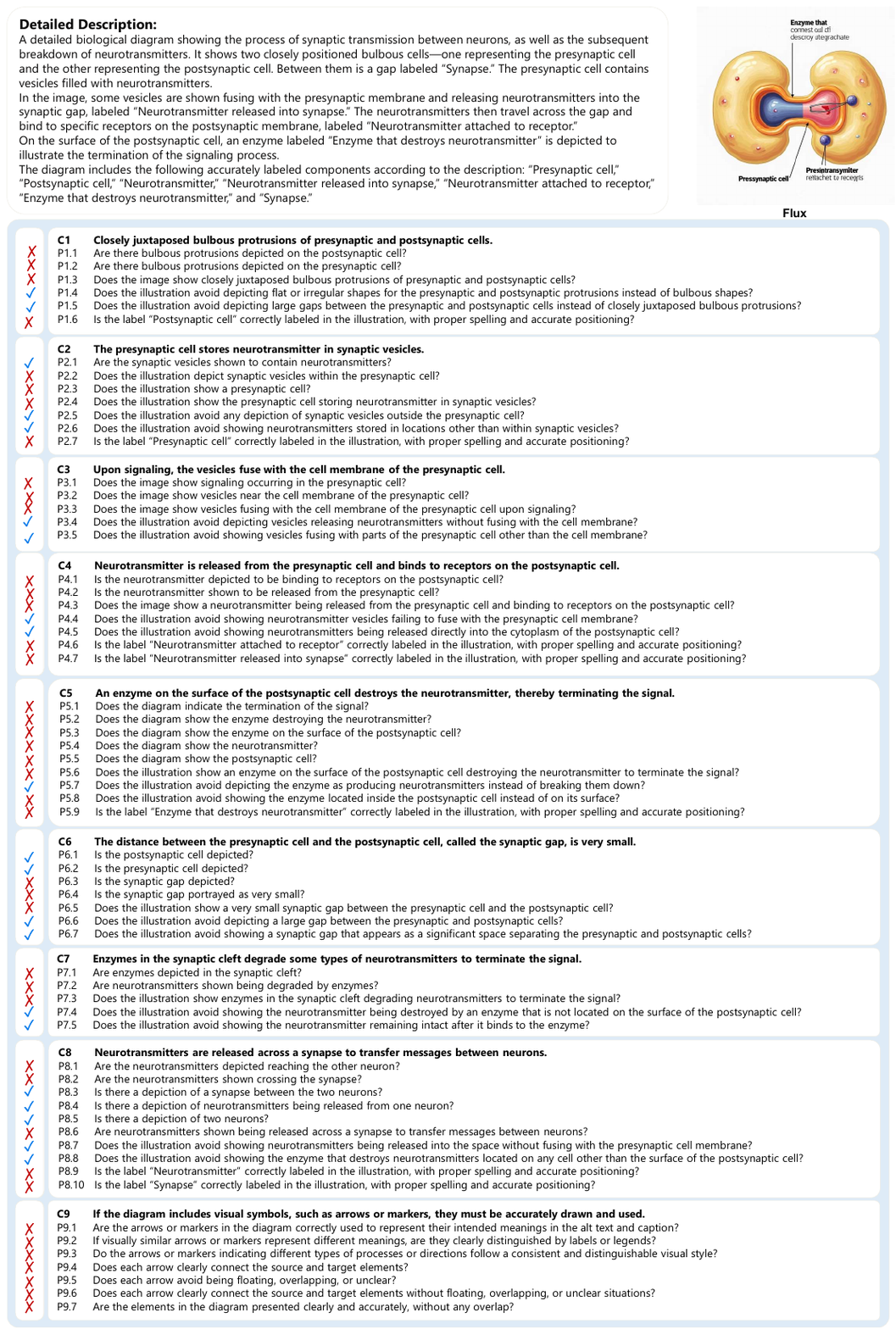}
    \caption{Evaluation case of Flux in the first task.}
    \label{fig:E1-3}
\end{figure*}

\begin{figure*}[h!]
    \centering
    \includegraphics[width=0.85\textwidth]{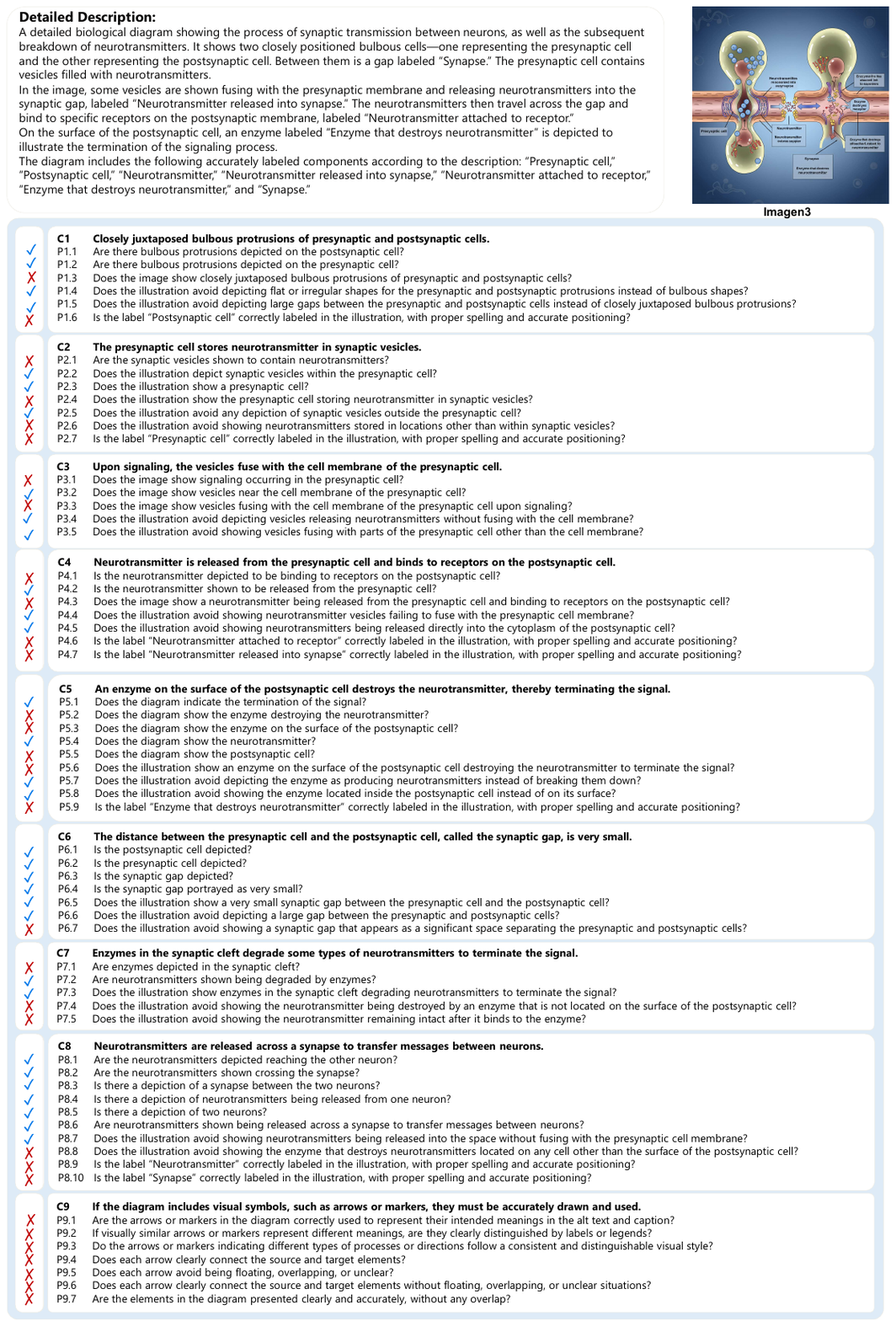}
    \caption{Evaluation case of Imagen3 in the first task.}
    \label{fig:E1-4}
\end{figure*}

\begin{figure*}[h!]
    \centering
    \includegraphics[width=0.85\textwidth]{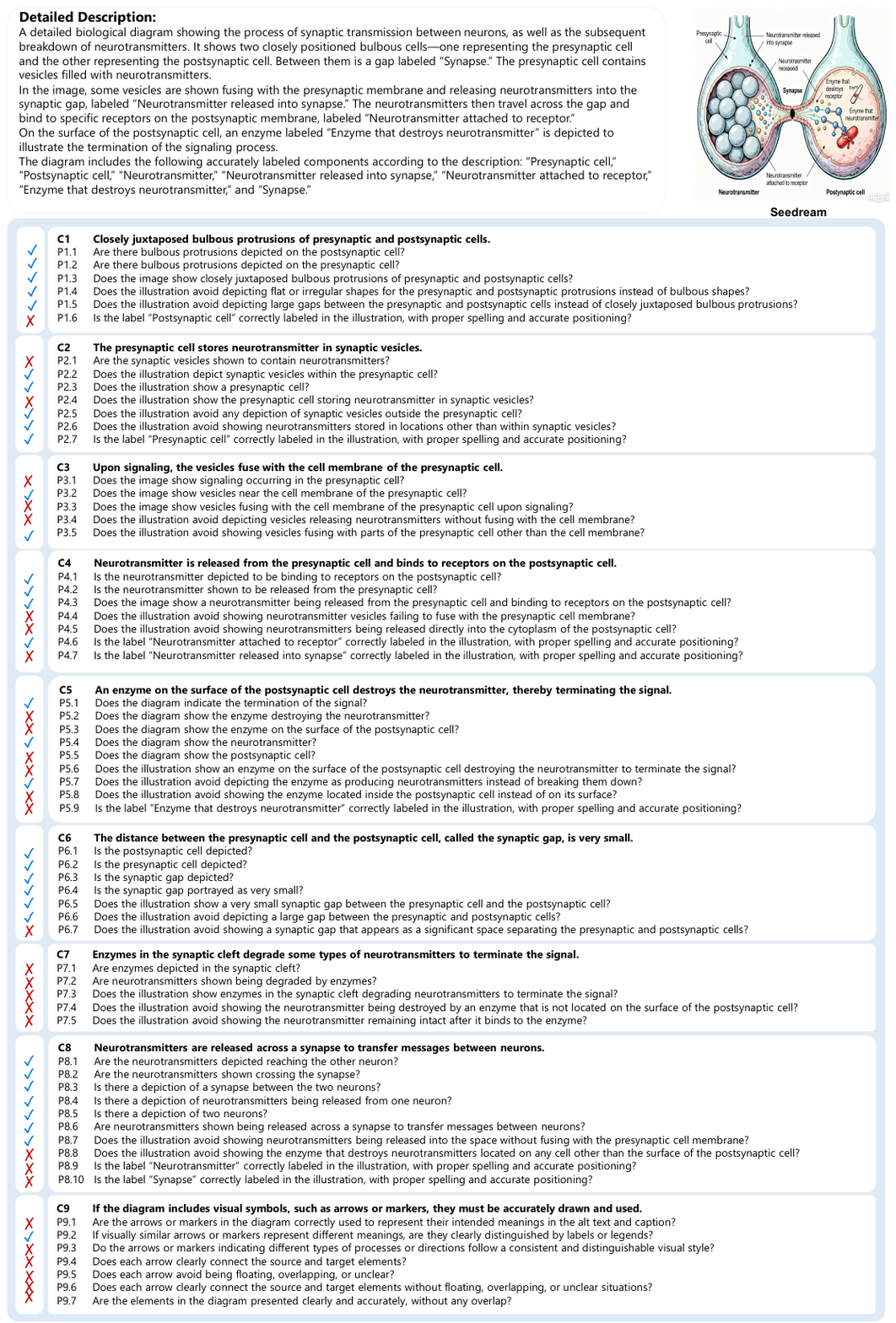}
    \caption{Evaluation case of Seedream in the first task.}
    \label{fig:E1-5}
\end{figure*}

\begin{figure*}[h!]
    \centering
    \includegraphics[width=0.85\textwidth]{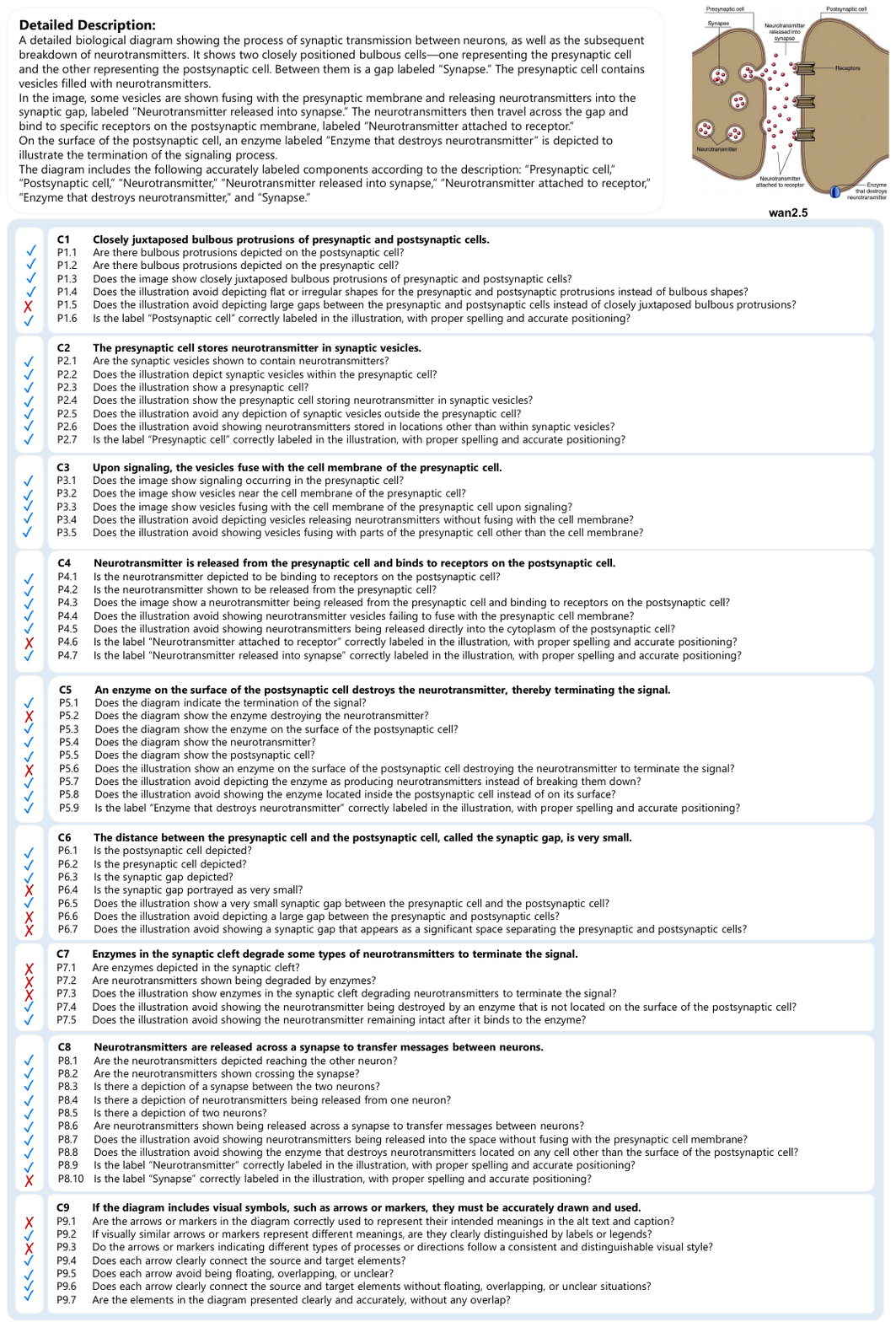}
    \caption{Evaluation case of Wan2.5 in the first task.}
    \label{fig:E1-6}
\end{figure*}

\begin{figure*}[h!]
    \centering
    \includegraphics[width=0.85\textwidth]{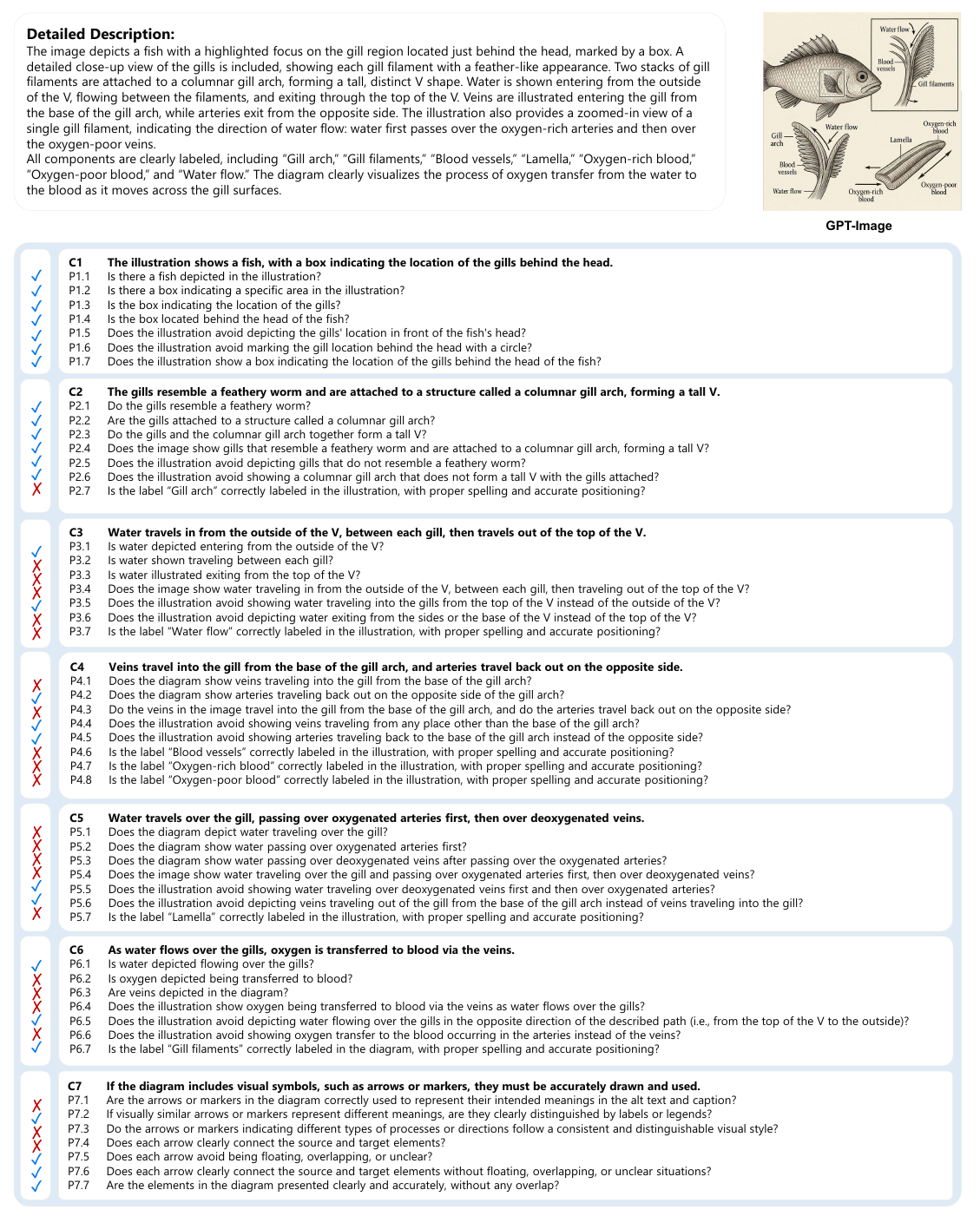}
    \caption{Evaluation case of GPT-4o in the second task.}
    \label{fig:E2-1}
\end{figure*}

\begin{figure*}[h!]
    \centering
    \includegraphics[width=0.85\textwidth]{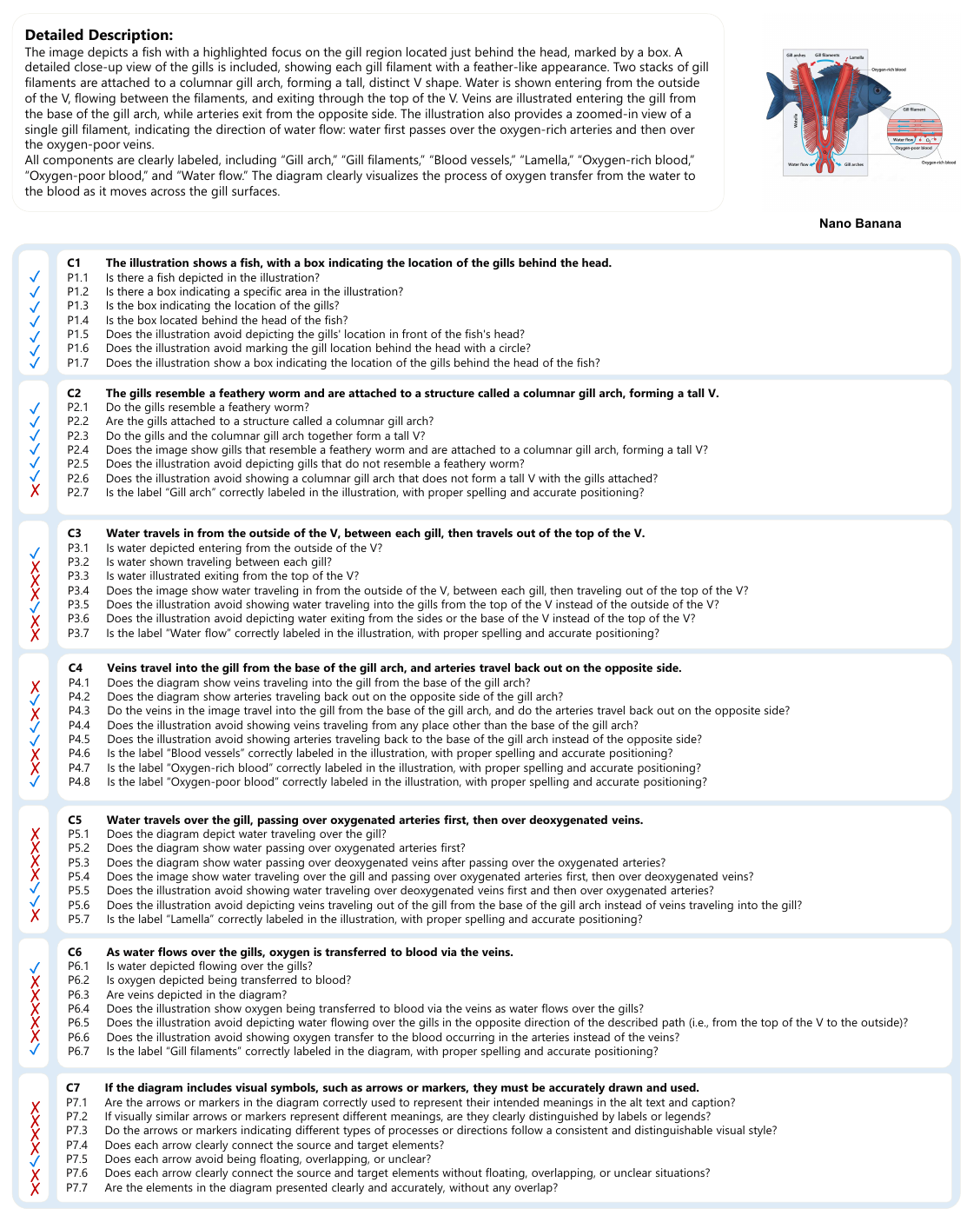}
    \caption{Evaluation case of Nano Banana in the second task.}
    \label{fig:E2-2}
\end{figure*}

\begin{figure*}[h!]
    \centering
    \includegraphics[width=0.85\textwidth]{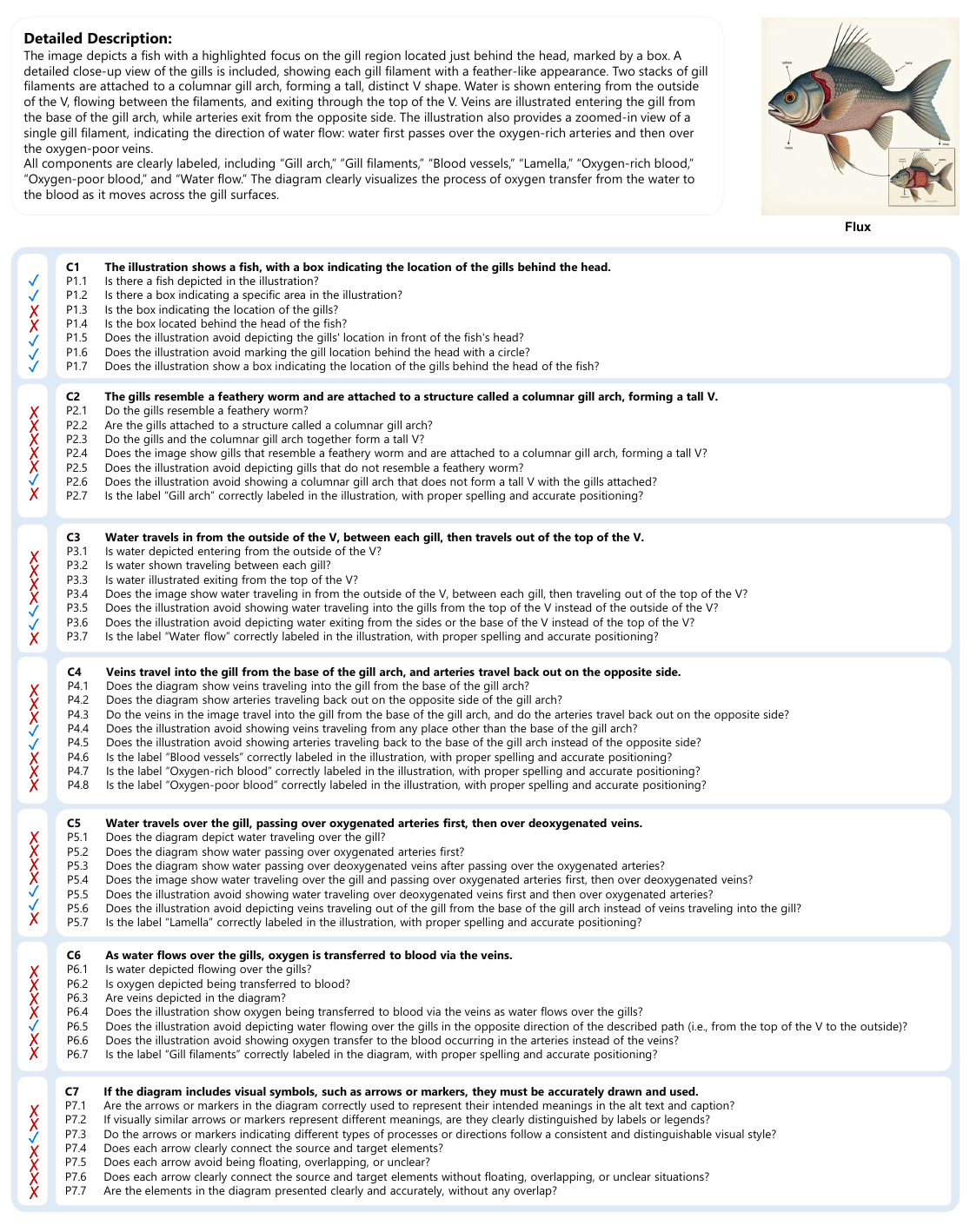}
    \caption{Evaluation case of Flux in the second task.}
    \label{fig:E2-3}
\end{figure*}

\begin{figure*}[h!]
    \centering
    \includegraphics[width=0.85\textwidth]{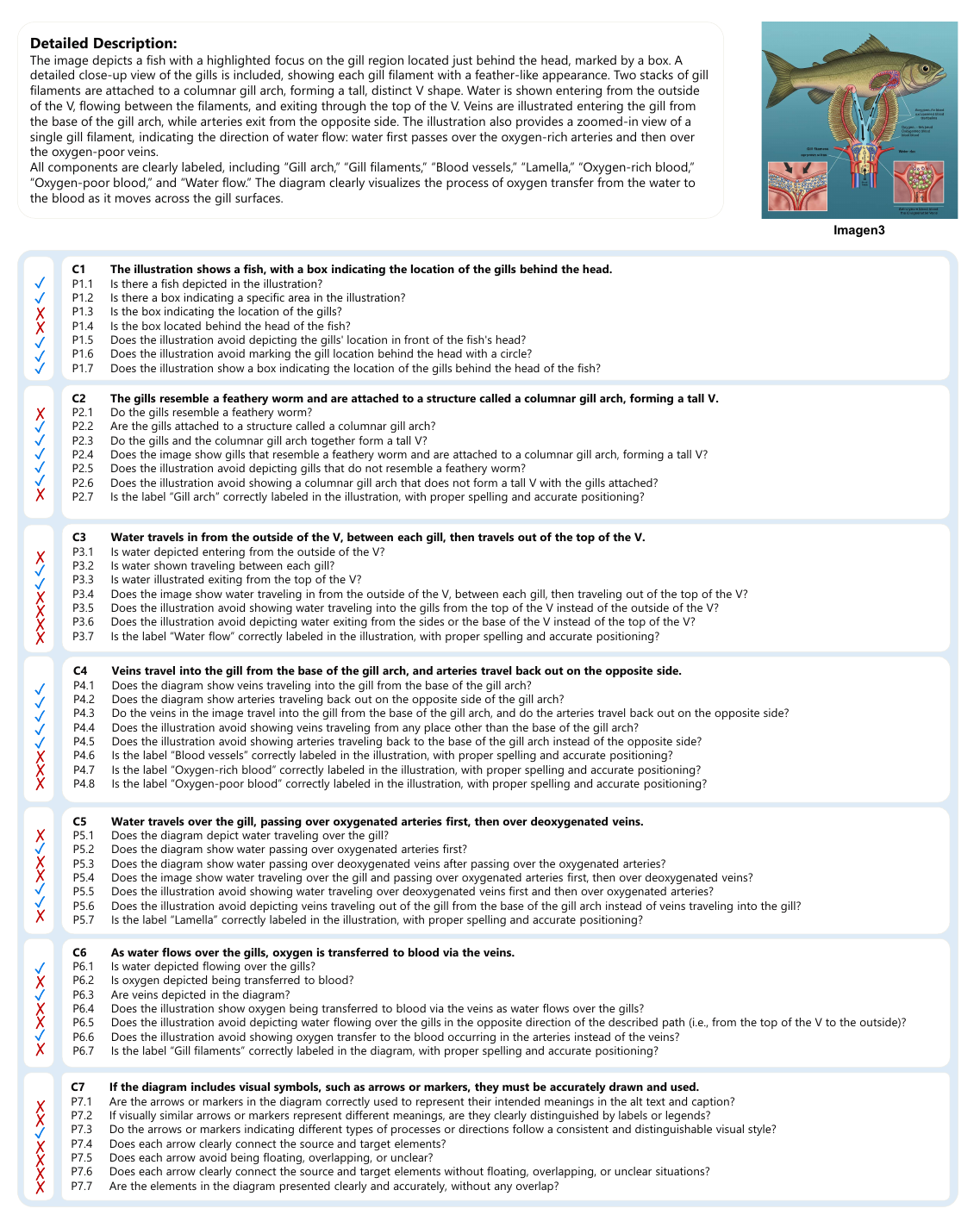}
    \caption{Evaluation case of Imagen3 in the second task.}
    \label{fig:E2-4}
\end{figure*}

\begin{figure*}[h!]
    \centering
    \includegraphics[width=0.85\textwidth]{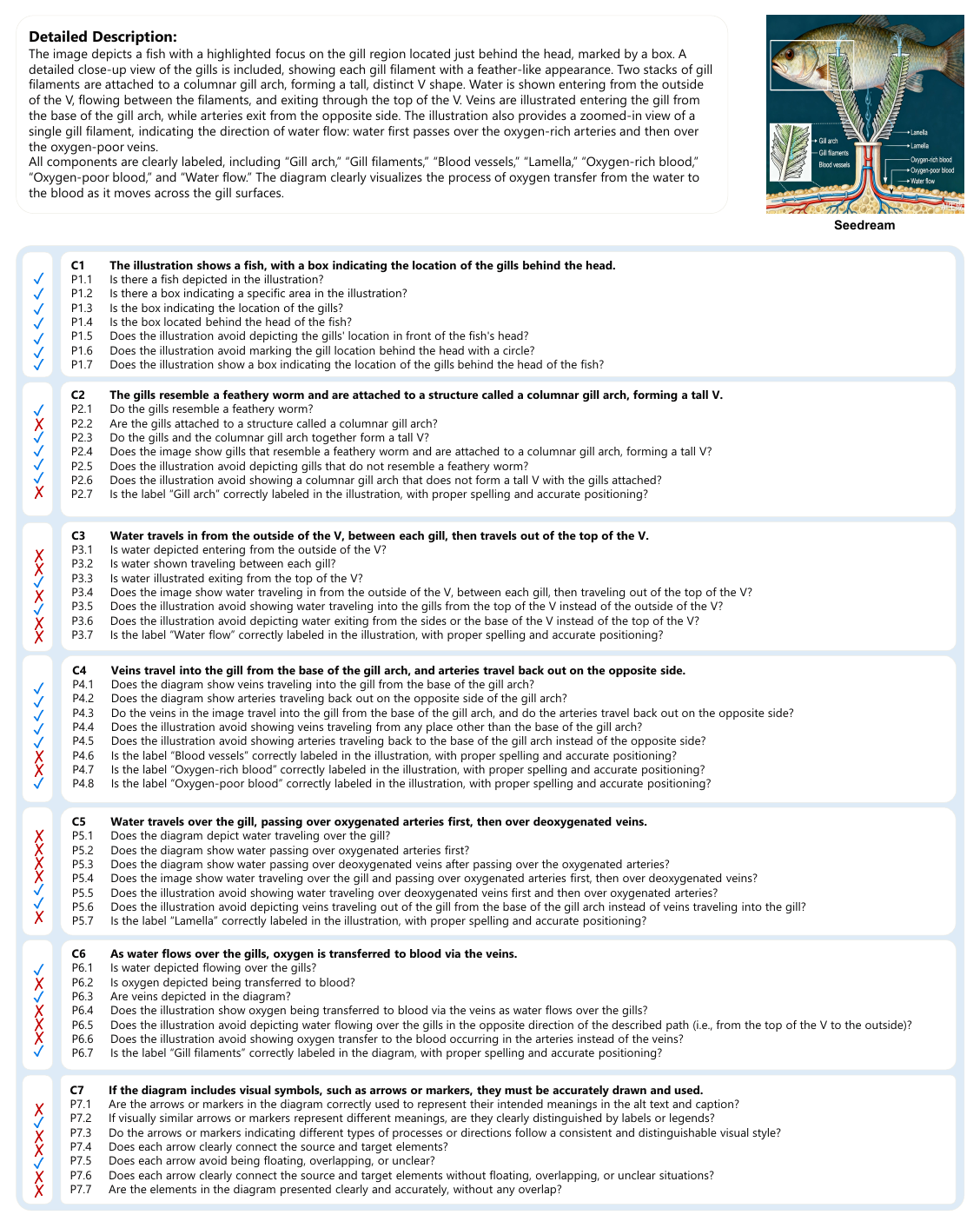}
    \caption{Evaluation case of Seedream in the second task.}
    \label{fig:E2-5}
\end{figure*}

\begin{figure*}[h!]
    \centering
    \includegraphics[width=0.85\textwidth]{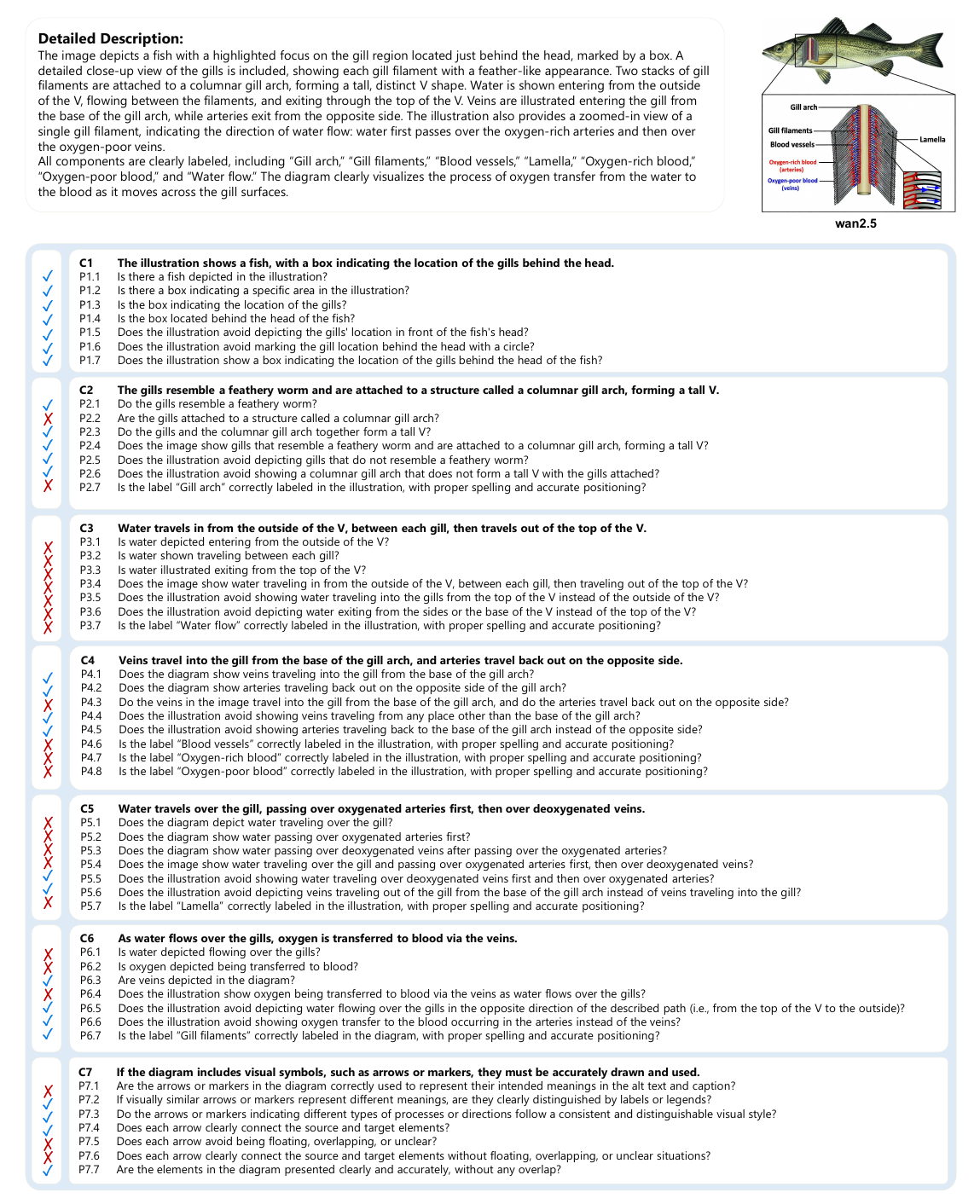}
    \caption{Evaluation case of Wan2.5 in the second task.}
    \label{fig:E2-6}
\end{figure*}

To better illustrate the evaluation of our benchmark, we sample two different tasks. The first task aims to generate the process of synaptic transmission between two neurons, as shown in Figures \ref{fig:E1-1}-\ref{fig:E1-6}. The second task aims to generate the gill structure of a fish, highlighting the gill arches, gill filaments, and the interactions between blood flow and water flow, as shown in Figures \ref{fig:E2-1}-\ref{fig:E2-6}.

Although six baseline models demonstrate strong generative capabilities in open-domain image tasks, they still face substantial challenges when producing professional scientific diagrams that are structurally complex, process-rich, and demand a high level of precision. They commonly exhibit issues such as missing processes, misaligned structures, and incorrect labeling, making it difficult for them to meet the rigorous requirements of high-quality technical image generation.

\section{Online Benchmark and Code Release}
\label{sec:release}

We released our online evaluation benchmark and code at \href{https://github.com/kodenii/ProImage-Bench}{GitHub}. Our TechImage-Bench dataset will be released under the MIT license.

\section{Ethics Statement}
\label{sec:ethical}

\paragraph{Inappropriate Content and Discrimination}

Considering the involvement of LMMs, TechImage-Bench places strong emphasis on mitigating risks related to inappropriate content and discrimination during its construction. All textual materials undergo manual review to ensure that no responsible-risk content is included in the benchmark.

\paragraph{Anti-misuse}

We notice the risk of technical image generation, which may also be misused to fabricate misleading figures, deceptive technical illustrations, or fake patent sketches. Therefore, we advocate for transparency in technical image generation, and we hope that TechImage-Bench can promote more rigorous, verifiable, and responsible research in this field.




\end{document}